\documentclass[12pt,letterpaper]{article}
\usepackage[a4paper, total={7in, 10in}]{geometry}

\usepackage{graphicx}
\usepackage{helvet}
\usepackage{authblk}
\usepackage{hyperref}
\usepackage{amsmath} 
\usepackage{amssymb} 
\usepackage{orcidlink} 
\usepackage{booktabs}
\usepackage{multirow}
\usepackage[super,comma,sort&compress]{natbib}
\usepackage{wrapfig}
\usepackage{caption}
\usepackage{pifont}
\makeatletter
\renewcommand{\maketitle}{\bgroup\setlength{\parindent}{0pt}
\begin{flushleft}
  \textbf{\@title}
  
  \@author
\end{flushleft}\egroup}
\makeatother
\usepackage[leftcaption]{sidecap}
\usepackage{subcaption}
\usepackage{xcolor}
\usepackage[normalem]{ulem}

\newcommand{\blue}[1]{\textcolor{blue}{#1}}
\newcommand{\orange}[1]{\textcolor{orange}{#1}}

\def\m{{\bf m}}

\def\0{{\bf 0}}
\def\1{{\bf 1}}

\def\argmin{\mathop{\rm argmin}}

\numberwithin{theorem}{section}
\numberwithin{lemma}{section}
\numberwithin{remark}{section}
\numberwithin{cor}{section}
\numberwithin{proposition}{section}
\numberwithin{definition}{section}

\newcommand{\tabref}[1]{Table~\ref{#1}}

\newcommand{\figref}[1]{Figure~\ref{#1}}

\newcommand{\eqnref}[1]{Eqn. (\ref{#1})}

\renewcommand{\tilde}{\widetilde}
\renewcommand{\hat}{\widehat}
\renewcommand{\frac}{\tfrac}

\newcommand{\pmv}[1]{\scriptsize$\pm$#1}
\newcommand{\cmark}{\ding{51}}%
\newcommand{\xmark}{\ding{55}}%
\usepackage{numcompress}
\title{CLEAR-HPV: Interpretable Concept Discovery for HPV-Associated Morphology in Whole-Slide Histology}
\date{}

\author[1,*,\orcidlink{0009-0001-9225-9188}]{Weiyi Qin}
\author[2]{Yingci Liu-Swetz}
\author[1]{Shiwei Tan}
\author[1,3,*]{Hao Wang}

\affil[1]{Department of Computer Science, Rutgers University, New Brunswick, NJ, USA}
\affil[2]{Rutgers Health, Rutgers University, Newark, NJ, USA}
\affil[3]{Lead contact}

\affil[*]{Correspondence: wq50@cs.rutgers.edu (W.Q.), hw488@cs.rutgers.edu (H.W.)}

\begin{document}

\maketitle

\section*{SUMMARY}

Human papillomavirus (HPV) status is a critical determinant of prognosis and treatment response in head and neck and cervical cancers. Although attention-based multiple instance learning (MIL) achieves strong slide-level prediction for HPV-related whole-slide histopathology, it provides limited morphologic interpretability. To address this limitation, we introduce Concept-Level Explainable Attention-guided Representation for HPV (CLEAR-HPV), a framework that restructures the MIL latent space using attention to enable concept discovery without requiring concept labels during training. Operating in an attention-weighted latent space, CLEAR-HPV automatically discovers keratinizing, basaloid, and stromal morphologic concepts, generates spatial concept maps, and represents each slide using a compact concept-fraction vector. CLEAR-HPV's concept-fraction vectors preserve the predictive information of the original MIL embeddings while reducing the high-dimensional feature space (e.g., 1536 dimensions) to only 10 interpretable concepts. CLEAR-HPV demonstrates consistent concept structure across TCGA-HNSCC, TCGA-CESC, and CPTAC-HNSCC, providing compact, concept-level interpretability through a general, backbone-agnostic framework for attention-based MIL models of whole-slide histopathology. All original code, preprocessing scripts, and trained model checkpoints are available on GitHub (\url{https://github.com/Wang-ML-Lab/CLEAR-HPV}))

% Attention-based multiple instance learning (MIL) enables strong slide-level prediction but offers limited insight into the underlying morphologic structure. We introduce CLEAR-HPV (Concept-Level Explainable Attention-guided Representation for HPV), a framework that restructures the MIL latent space using attention to enable annotation-free concept discovery. Operating in a attention-weighted latent space, CLEAR-HPV identifies coherent keratinizing, basaloid, and stromal morphologic concepts, generates spatial concept maps, and constructs compact concept-fraction slide representations. These concept-fraction representations retain the predictive information of the original MIL embeddings, as shown by comparable performance when used as inputs to downstream classifiers. Applied across TCGA-HNSCC, TCGA-CESC, and CPTAC-HNSCC, CLEAR-HPV demonstrates consistent cross-cohort reproducibility, indicating that the discovered concepts represent durable HPV-associated morphology rather than dataset-specific artifacts. CLEAR-HPV is computationally efficient and provides biologically grounded, concept-level interpretability for whole-slide histopathology independent of classification accuracy.

%These concept-fraction representations preserve the discriminative structure encoded in the original MIL embeddings, enabling downstream classifiers to recover comparable slide-level predictions.

\section*{KEYWORDS}

%%%  Include up to 10 keywords, separated by commas. 
%%%  Keywords entered in EM are not carried over; only 
%%%  keywords included in the main text will be used 
%%%  in the final article metadata. Please note that 
%%%  for some journals, keywords are chosen by the editors.

computational pathology, multiple instance learning, interpretable deep learning, concept discovery, head and neck cancer, human papillomavirus (HPV), whole slide image analysis

\section*{INTRODUCTION}

% Human papillomavirus (HPV) status plays a critical role in shaping prognosis and guiding therapeutic decisions for head and neck and cervical squamous cell carcinoma. 
HPV-associated head and neck and cervical cancers together account for 690,000 new cases worldwide each year\cite{de2020global}, with HPV status strongly stratifying survival, treatment intensity, and long-term functional outcomes, making accurate and interpretable HPV assessment a problem of major clinical and public health significance. 
HPV-positive tumors are often non-keratinizing or basaloid\cite{Ang2010NEJM, Gillison2000JNCI, Marur2008HNCChangingEpidemiology}, whereas HPV-negative tumors more commonly display keratinizing squamous morphology\cite{Ang2010NEJM, Gillison2000JNCI, Linton2013BasaloidSCC}. 
% However, substantial morphologic overlap and variability mean that HPV status cannot be reliably inferred from histology alone\cite{Shah2014SCCVariantsReview}, and ancillary immunohistochemical or molecular assays remain the standard of care\cite{Kather2019MSI, Lewis2018HPVTestingGuideline}. 
However, substantial morphologic overlap and variability mean that HPV status cannot be reliably inferred by human observers from routine histologic assessment alone\cite{Shah2014SCCVariantsReview}; ancillary immunohistochemical or molecular assays therefore remain the standard of care\cite{Kather2019MSI, Lewis2018HPVTestingGuideline}, but they often incur substantially higher costs, owing to additional reagents, instrumentation, and technical processing.

On the other hand, digital histology (and histopathology) and the use of whole slide images (WSIs) have recently shown promising accuracy in capturing nuanced patterns that human observers often miss; however, they are often lacking in interpretability, limiting their clinical adoption.
% Despite such limitations, digital histopathology and the use of whole slide images (WSIs) for HPV prediction become more widespread, there is a pressing need for computational methods that not only classify slides accurately but also reveal how predictions relate to recognizable, biologically grounded morphologic patterns, with stability across different staining, scanning, and institutional settings.
% As digital pathology and the use of whole slide images (WSIs) for HPV prediction become more widespread, there is a pressing need for computational methods that not only classify slides accurately but also reveal how predictions relate to recognizable, biologically grounded morphologic patterns, with stability across different staining, scanning, and institutional settings.
% In the past decade, digital pathology and the use of whole slide images (WSIs) for HPV prediction have become more widespread, 
As a result, there is a pressing need for the best of both worlds: computational histology methods that not only classify slides accurately but also reveal how predictions relate to recognizable, biologically grounded morphologic patterns, with stability across different staining, scanning, and institutional settings.

% These clinical and diagnostic challenges have motivated the development of deep learning methods, including recent vision foundation models, for automated HPV prediction from WSIs. 
Recent deep learning methods, including vision foundation models, have achieved strong performance in WSI classification across diverse diagnostic and molecular tasks \cite{Campanella2019Clinical, Coudray2018NSCLC, Liu2020_AI_NodalMetastasisDetection, Kather2019MSI}, but most models function as black boxes that offer limited interpretability on what histologic patterns are driving their predictions \cite{samek2017explainable, arrieta2020explainable, tjoa2020survey}. Weakly supervised multiple instance learning (MIL) is a widely adopted framework for WSI analysis, and popular MIL models (e.g., ABMIL \cite{ITW:2018}, CLAM \cite{lu2021clam}, TransMIL \cite{shao2021transmil}) treat each slide as a large, heterogeneous collection of tiles with only slide-level labels. Widely used interpretability methods such as attention heatmaps and Grad-CAM\cite{Selvaraju_2019} indicate where a model attends, but do not provide concept-level, human-understandable explanations of what histologic patterns are driving its predictions. As a result, current approaches offer only coarse, qualitative cues and cannot identify the discrete, reproducible morphologic concepts present in WSIs. This limits biological insight, reduces reproducibility across sites, and weakens trust in model predictions, therefore limiting clinical deployment.

These interpretability limitations motivate a closer examination of how MIL models internally encode morphology to determine whether their representations can be reorganized into \emph{clinically meaningful and human-understandable concepts}. Prior work has shown that deep neural networks naturally organize intermediate features into latent spaces that encode semantic or visual factors \cite{bengio2014representationlearningreviewnew, caron2021emergingpropertiesselfsupervisedvision}. In attention-based MIL models, the tile-level embeddings produced before attention pooling define an intermediate $h$-space that captures the morphologic features learned by the model across tumor and stromal regions. Latent spaces such as the $h$-space can be reorganized into human-interpretable concepts \cite{koh2020conceptbottleneckmodels}, and concept discovery from neural embeddings has been shown to recover coherent visual structures \cite{goyal2020explainingclassifierscausalconcept, ghorbani2019automaticconceptbasedexplanations,9359803,xia-etal-2025-hgclip,gu2025radalignadvancingradiologyreport}. Together, these observations suggest that MIL backbones already encode rich morphologic structure, but require an attention-aware organization strategy to make this structure explicit and biologically interpretable.

In this work, we show that attention mechanisms in MIL induce a latent morphologic structure that can be reorganized into discrete histologic concepts without tile-level annotations. We develop CLEAR-HPV (Concept-Level Explainable Attention-guided Representation for HPV), a framework that restructures the attention-weighted MIL $h$-space to enable \emph{annotation-free concept discovery} in HPV-related histopathology. Rather than modifying the classifier or explicitly optimizing for higher accuracy, CLEAR-HPV operates \emph{post hoc} on the latent embeddings of a \emph{trained} model (e.g., CLAM\cite{lu2021clam}), using attention weights to focus concept discovery on tiles the model already considers informative. The framework yields coherent keratinizing, basaloid, and stromal morphologic concepts that align with established HPV-associated patterns, together with two complementary interpretable outputs, \emph{spatial concept maps} that show where concepts appear across each slide, and \emph{compact concept-fraction representations} that summarize slide-level tissue composition in a quantitative, low-dimensional form.

\begin{figure}[!t]
  \centering
  \includegraphics[width=\linewidth]{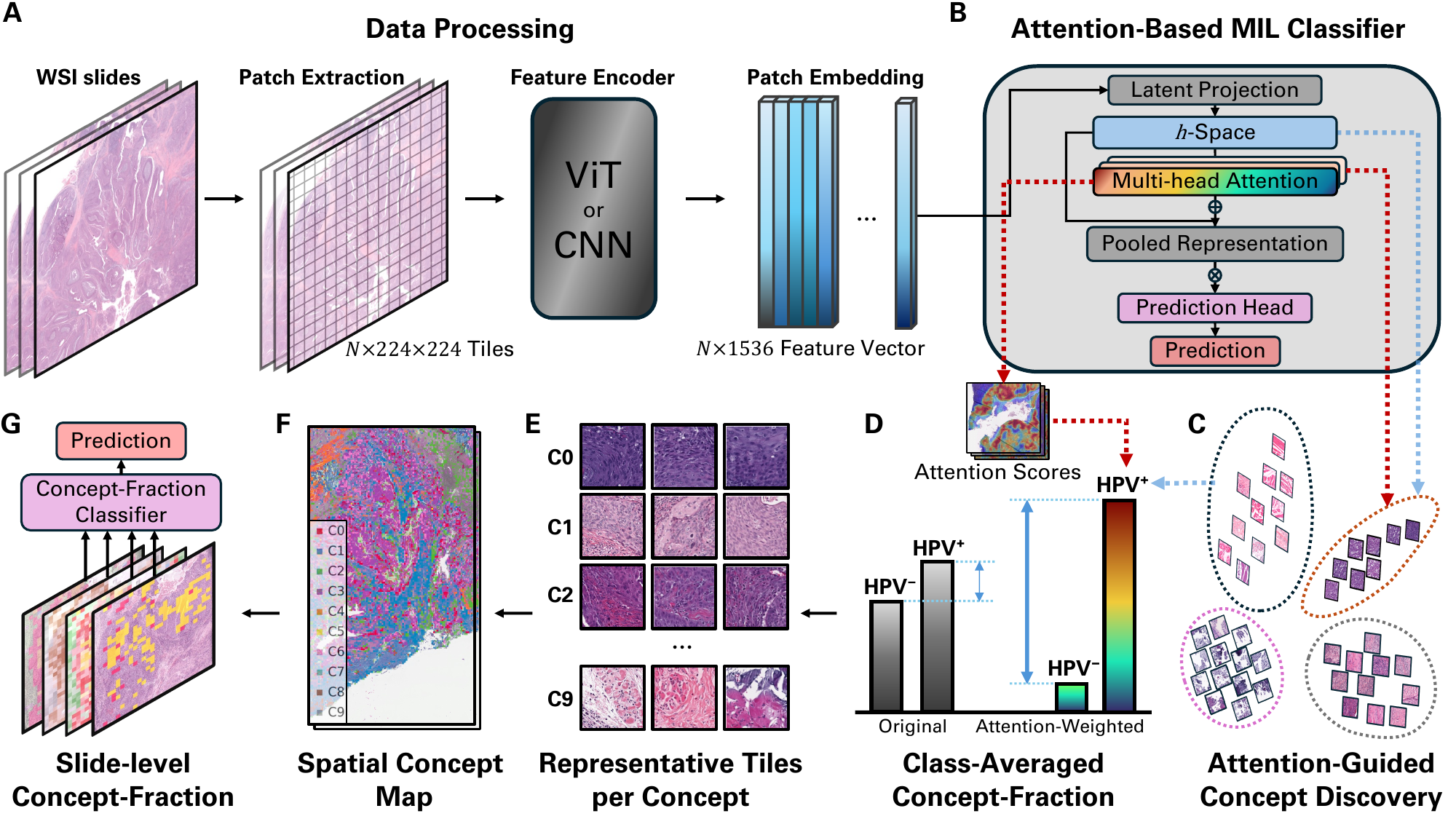}
  \caption{
    \textbf{Overview of the CLEAR-HPV framework.}
    {\textbf{(A)} Data processing pipeline: WSIs are decomposed into fixed-size tiles, encoded with a pretrained ViT or CNN, and converted into patch-level feature embeddings. 
    \textbf{(B)} An attention-based MIL classifier projects embeddings into the $h$-space latent representation and uses multi-head attention to compute tile-level contributions, which are pooled into a single slide-level embedding for HPV prediction. 
    \textbf{(C)} CLEAR-HPV performs annotation-free concept discovery on attention-weighted $h$-space embeddings to identify coherent morphologic concepts. 
    % \textbf{(D)} Class-averaged concept-fraction vectors for HPV-positive and HPV-negative cohorts, obtained by averaging slide-level concept fractions within each group. Attention-weighted concept fractions highlight clearer differences between the two groups.
    \textbf{(D)} Using the discovered concepts, each slide is represented by a concept-fraction vector, which is then averaged across slides to obtain class-averaged concept-fraction vectors that summarize morphologic composition for HPV-positive and HPV-negative cohorts.
    \textbf{(E)} Representative tiles illustrate the characteristic morphology captured by each discovered concept. 
    \textbf{(F)} Spatial concept maps visualize the distribution of concepts across the WSIs, revealing their spatial organization. 
    \textbf{(G)} Slide-level concept-fraction vectors provide an interpretable representation that supports a concept-fraction classifier, which recovers MIL predictive performance while offering concept-level explanations. More details are available in the Methods section.}
    } 
  \label{fig:main}
\end{figure}

Applied to three cohorts of data, TCGA-HNSCC \cite{weinstein2013cancer}, TCGA-CESC \cite{weinstein2013cancer}, and CPTAC-HNSCC \cite{edwards2015cptac}, CLEAR-HPV discovers stable concepts and consistent concept-fraction patterns that generalize across cohorts, indicating that the discovered morphology reflects consistent HPV-associated structure rather than dataset-specific artifacts. The resulting concept-fraction representations preserve the discriminative structure encoded in the original MIL embeddings, allowing downstream classifiers to recover comparable slide-level predictions while operating on an interpretable concept space --- achieving the best of both worlds. Together, these findings demonstrate that the latent feature space of attention-based MIL already contains rich, biologically meaningful organization, and that our attention-guided concept discovery can expose this structure without sacrificing predictive performance. \figref{fig:main} shows an overview of our CLEAR-HPV; implementation details are described in the Methods section.

In summary, our primary contributions are as follows:
\begin{enumerate}

\item We introduce CLEAR-HPV, the first general framework to automatically discover pathology-relevant morphologic concepts for HPV prediction without tile-level supervision (or annotation). 
\item We demonstrate that CLEAR-HPV leverages the attention-weighted latent space in a deep learning model to produce spatial concept maps and concept-fraction vectors, 
offering biologically grounded, concept-level interpretability for whole-slide histopathology. 
\item 
CLEAR-HPV preserves the predictive performance of the interpreted model while reducing its high-dimensional features (e.g., 1536 dimensions) to only 10 interpretable concepts. 
{\item We show that CLEAR-HPV is compatible with diverse attention-based MIL backbones and retains strong slide-level performance across diverse architectural designs, enabling robust and consistent concept discovery beyond a single model instantiation.}
\item We further show that these concept-level representations are stable across different cohorts (e.g., TCGA and CPTAC), preserve clinically relevant predictive signal under transfer, and reveal cross-cohort consistency of HPV-related morphology.

\end{enumerate}

\section*{RESULTS} 

% example:
% https://www.cell.com/action/showPdf?pii=S2666-3899%2825%2900274-0
% https://www.cell.com/action/showPdf?pii=S2666-3899%2825%2900275-2
In this study, we evaluated CLEAR-HPV across three independent WSI cohorts (datasets) to examine whether biologically coherent and interpretable concepts can be discovered across diverse clinical and technical settings. The TCGA-HNSCC cohort included 102 patients and 106 diagnostic WSIs (38 HPV-positive, 64 HPV-negative). The TCGA-CESC cohort consisted of 146 patients and 154 WSIs, predominantly HPV-positive (138 HPV-positive, 8 HPV-negative)\cite{weinstein2013cancer}. The CPTAC-HNSCC dataset contributed 112 HPV-negative patients and 368 WSIs, serving as an external validation cohort collected under a different study protocol\cite{edwards2015cptac}. These datasets differ substantially in HPV prevalence, staining and scanning protocols, and clinical outcomes, allowing us to assess whether CLEAR-HPV identifies stable morphologic concepts rather than cohort-specific artifacts. 

CLEAR-HPV is designed as a post-hoc explainability framework; its goal is \emph{not} to improve predictive accuracy but to discover and interpret the morphologic concepts encoded in the model’s internal representations. 
Compared to the interpreted (explained) deep learning model that uses \emph{high-dimensional} (e.g., 1536 dimensions), \emph{uninterpretable} embeddings, CLEAR-HPV discovers a \emph{compact}, \emph{interpretable} set of concepts, e.g., only 10 concepts. % that can achieve similar accuracy. 
Therefore, results are considered \emph{very strong} as long as \emph{comparable} predictive performance (e.g., AUC, ACC, F1) can be achieved using CLEAR-HPV's discovered concepts.

We use CLAM\cite{lu2021clam}, a widely adopted attention-based multiple instance learning (MIL) method, as the primary target backbone (base) model to explain. CLAM provides tile-level attention scores and a learned intermediate latent space (the $h$-space). Together, these form the foundation for concept discovery. We also provide results on {three} other backbone models and their variants to demonstrate the generality of CLEAR-HPV  (\tabref{tab:backbone_main}).

A consistent 10-fold protocol was applied to each cohort, enabling systematic assessment of interpretability and robustness across heterogeneous datasets (\figref{fig:main}).

\subsection*{Base model performance}

\tabref{tab:clam_crosscohort_main} shows that, on TCGA-HNSCC, the CLAM attention-based MIL backbone achieved consistent slide-level performance (ACC = 0.77 $\pm$ 0.06, AUC = 0.86 $\pm$ 0.05), indicating that its learned representations capture generalizable, discriminative structure. 

We then evaluated the model in a fully zero-shot setting on external cohorts \emph{without any calibration or retraining}. On TCGA-CESC, a cohort dominated by HPV-positive tumors, performance decreased as expected due to differences in tissue origin and histomorphologic context (AUC $\approx$ 0.68), but the model retained very high precision ($\approx$ 0.98), indicating preservation of class-specific structure under domain shift. On CPTAC-HNSCC, where only HPV-negative cases are available for evaluation, accuracy remained consistent (0.70 $\pm$ 0.12), demonstrating robustness to moderate staining and scanner variability. Together, these zero-shot evaluations show that the CLAM backbone maintains consistent decision behavior across cohorts and preserves transferable class-related structure in its latent representation, making it suitable for downstream concept-level analysis.

\begin{table}[!t]
\centering
\footnotesize
\setlength{\tabcolsep}{10pt}
\renewcommand{\arraystretch}{1.15}

\begin{tabular}{lccccc}
\hline
\textbf{Dataset} & \textbf{ACC} & \textbf{AUC} & \textbf{Prec} & \textbf{Rec} & \textbf{F1} \\
\hline

TCGA-HNSCC 
 & 0.765\pmv0.063 
 & 0.863\pmv0.051 
 & 0.788\pmv0.098
 & 0.673\pmv0.145
 & 0.696\pmv0.101 \\

TCGA-CESC 
 & 0.593\pmv0.083 
 & 0.684\pmv0.154 
 & 0.977\pmv0.029
 & 0.581\pmv0.176
 & 0.716\pmv0.073 \\

CPTAC-HNSCC\textsuperscript{a}
 & 0.704\pmv0.120 
 & N/A 
 & N/A 
 & N/A 
 & N/A \\
\hline
\end{tabular}

\vspace{4pt}
    \caption{
    \textbf{Cross-cohort generalization of the baseline CLAM model.}
    }
\vspace{2pt}

\textsuperscript{a}\,AUC, Precision, Recall, and F1 are not applicable (N/A) because the CPTAC-HNSCC cohort contains only HPV-negative slides (single-class).

\label{tab:clam_crosscohort_main}
\end{table}

% \subsection*{Concept discovery in $h$-space without needing label annotation}
\subsection*{Annotation-free concept discovery in the $h$-space}
%Good idea!

CLEAR-HPV is motivated by the observation that the latent $h$-space of attention-based MIL models contains rich morphologic structure that can be made explicit via concept discovery. Here, the $h$-space consists of tile-level embeddings produced before attention pooling, illustrated as a blue box in \figref{fig:main}(B). The attention weights learned by the backbone identify which regions contribute most to the final prediction, and attention pooling is typically used to pool these tile-level embeddings into a single embedding vector (i.e., ``Pooled Representation'' in~\figref{fig:main}(B)), which is fed into a prediction head for classification. CLEAR-HPV constructs an attention-guided representation by weighting each embedding $h_i$ by its attention score $\alpha_i$, thereby emphasizing diagnostically informative tiles while reducing background variation. More details are provided in the Methods section.

\paragraph{Concept-discovery methods and baselines.} To evaluate the discovered concepts, all concept discovery methods were applied to the TCGA-HNSCC training folds, which provide sufficient morphologic diversity for assessing cluster structure. Concept discovery was performed with $K=10$ concepts. This choice is supported by consistent empirical evidence: Table S1 shows that predictive performance remains stable across $K \in \{5,10,15\}$ with overlapping confidence intervals across all major metrics, indicating that $K=10$ achieves comparable performance without sensitivity to the exact choice of $K$. Table S2 further demonstrates that concept geometry is highly stable across resolutions, with forward persistence and reverse fragmentation both exceeding $0.96$, confirming that the discovered concepts are preserved under changes in $K$. The elbow analysis (Figure S1) provides additional support, showing diminishing returns in clustering compactness beyond $K=10$. We evaluate two variants of CLEAR-HPV: concepts produced from raw-$h$-space embeddings, i.e., ``CLEAR-HPV (raw-$h$)'', and concepts derived from attention-weighted (AW) $h$-space embeddings, i.e., ``CLEAR-HPV (AW-$h$)''. These form the primary concept sets used throughout the analysis. 
{For comparison, we evaluated several baseline methods that differ in where and how morphologic structure is extracted. Specifically, we considered (more details in the Methods section):} 
(1) heatmap-based grouping, which reflects which tiles the model attends to, without defining discrete concept distributions, 
(2) encoder-feature clustering, which generates unconstrained concept distributions derived from encoder feature representations, 
and (3) a Dirichlet concept model, which represents concept membership probabilistically and allows tiles and slides to express varying degrees of concept mixing. Together, these baselines compare different input representations and concept construction approaches, highlighting how attention-structured $h$-space representations influence the coherence and interpretability of discovered morphologic concepts.

\paragraph{Quantitative comparison of concept-discovery methods.} After defining the set of concept-discovery methods, we evaluate all approaches under a common framework to assess how well the resulting concepts summarize slide-level morphology and preserve diagnostically relevant signals. With $K$ concepts in total, each slide is summarized by a $K$-dimensional \emph{concept-fraction vector}
% (details in the Methods section) 
that quantifies the proportion of tiles assigned to each concept. The concept-fraction vector is the core representation used throughout our analysis, providing a unified and interpretable summary of slide-level morphology derived from tile-level concepts. To assess whether these concepts capture diagnostically meaningful information, we introduce a \emph{concept-fraction classifier} that maps concept fraction vectors to HPV status, without introducing additional trainable parameters. \begin{wrapfigure}{r}{0.4\linewidth}
\vskip -0.3cm
  \centering
  \includegraphics[width=\linewidth]{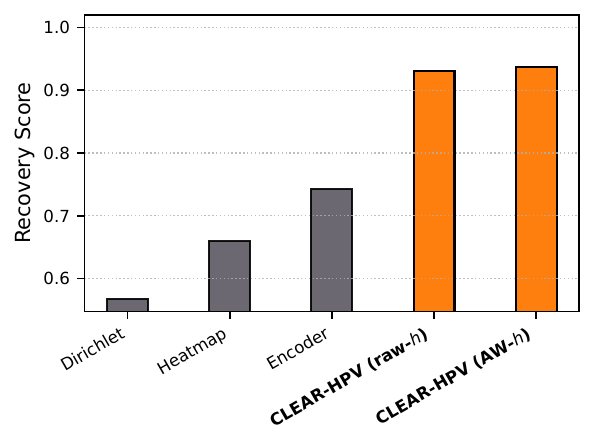}
  \caption{\textbf{Recovery score relative to the interpreted MIL model (CLAM) across ACC, AUC, F1, Precision, Recall (i.e., sensitivity), and Specificity.} For each method, the Euclidean distance $d$ between its metric vector (i.e., concatenation of Accuracy, AUC, etc.) and the interpreted model's is computed and converted to a similarity score $s = \frac{1}{1 + d}$. Higher scores indicate closer agreement with CLAM.
  }
  \label{fig:compare}
\end{wrapfigure} A detailed description is provided in the Methods section. We use common classification metrics such as Accuracy (ACC), Area Under the Curve (AUC), Precision, Recall, and F1 measure how well each representation preserves the predictive signal present in the original MIL embeddings. Note that our goal is \emph{not} to improve accuracy upon the original MIL backbone (e.g., CLAM). Instead, we aim to measure how well the concept-based explanations can retain the predictive performance of the original MIL backbone for HPV status. Therefore we consider these results strong when the CLEAR-HPV predictor achieves performance (e.g., ACC or AUC) comparable to the original MIL backbone.

\begin{table}[!t]
\centering
\footnotesize
\setlength{\tabcolsep}{10pt}
\renewcommand{\arraystretch}{1.15}

\begin{tabular}{lccccc}
\hline
\textbf{Method} & \textbf{ACC} & \textbf{AUC} & \textbf{Prec} & \textbf{Rec} & \textbf{F1} \\

Heatmap
 & 0.538\pmv0.110
 & 0.674\pmv0.147
 & 0.503\pmv0.137
 & 0.606\pmv0.215
 & 0.538\pmv0.114 \\

Dirichlet Concepts
 & 0.646\pmv0.071
 & 0.641\pmv0.134
 & 0.475\pmv0.314
 & 0.200\pmv0.208
 & 0.266\pmv0.193 \\

% Sinkhorn Concepts
%  & 0.569\pmv0.031
%  & 0.500\pmv0.000
%  & 0.000\pmv0.000
%  & 0.000\pmv0.000
%  & 0.000\pmv0.000 \\

Encoder Concepts
 & 0.709\pmv0.126
 & 0.847\pmv0.139
 & 0.648\pmv0.162
 & 0.835\pmv0.157
 & 0.714\pmv0.109 \\
\hline
\textbf{CLEAR-HPV (AW-$h$)}
 & 0.784\pmv0.141
 & 0.843\pmv0.117
 & 0.797\pmv0.197
 & 0.623\pmv0.284
 & 0.715\pmv0.206 \\

\textbf{CLEAR-HPV (raw-$h$)}
 & 0.749\pmv0.119
 & 0.889\pmv0.078
 & 0.770\pmv0.181
 & 0.633\pmv0.204
 & 0.684\pmv0.146 \\
\hline
\end{tabular}

\caption{
\textbf{Comparison of concept-discovery methods on TCGA-HNSCC.}
The heatmap provides an attention-only reference and does not define discrete concepts. Encoder concepts derive clusters directly from encoder feature representations. Within the $h$-space, Dirichlet concepts serve as an alternative concept discovery baseline applied to the unweighted latent space. In contrast, CLEAR-HPV performs concept discovery directly in the attention-weighted latent $h$-space. Metrics assess whether the resulting concept-fractions vectors retain the predictive signal present in the original MIL backbone. {Results on more metrics are provided in Table S3}.
}
\label{tab:hnscctrain_hnscc_test_cluster_main}

\end{table}

\tabref{tab:hnscctrain_hnscc_test_cluster_main} shows the performance of classifying HPV for all concept-discovery methods using only the discovered concept-fraction vectors, evaluated with a simple rule-based concept-fraction classifier. Our CLEAR-HPV's two variants, raw-$h$ and AW-$h$, capture complementary structure. Raw-$h$ preserves the intrinsic latent space learned by the MIL encoder and achieves the highest AUC, while AW-$h$ produces more coherent and stable morphologic concepts by amplifying high-attention tiles. Notably, even using only the discovered concept-fraction vector (with only $K=10$ dimensions), without access to the original tile embeddings (with 1536 dimensions), both variants retain predictive performance comparable to the CLAM backbone, indicating that the concept-fraction representation preserves the discriminative signal of the original model despite substantial dimensionality reduction.

\begin{figure}[!t]
  \centering
  \includegraphics[width=\linewidth]{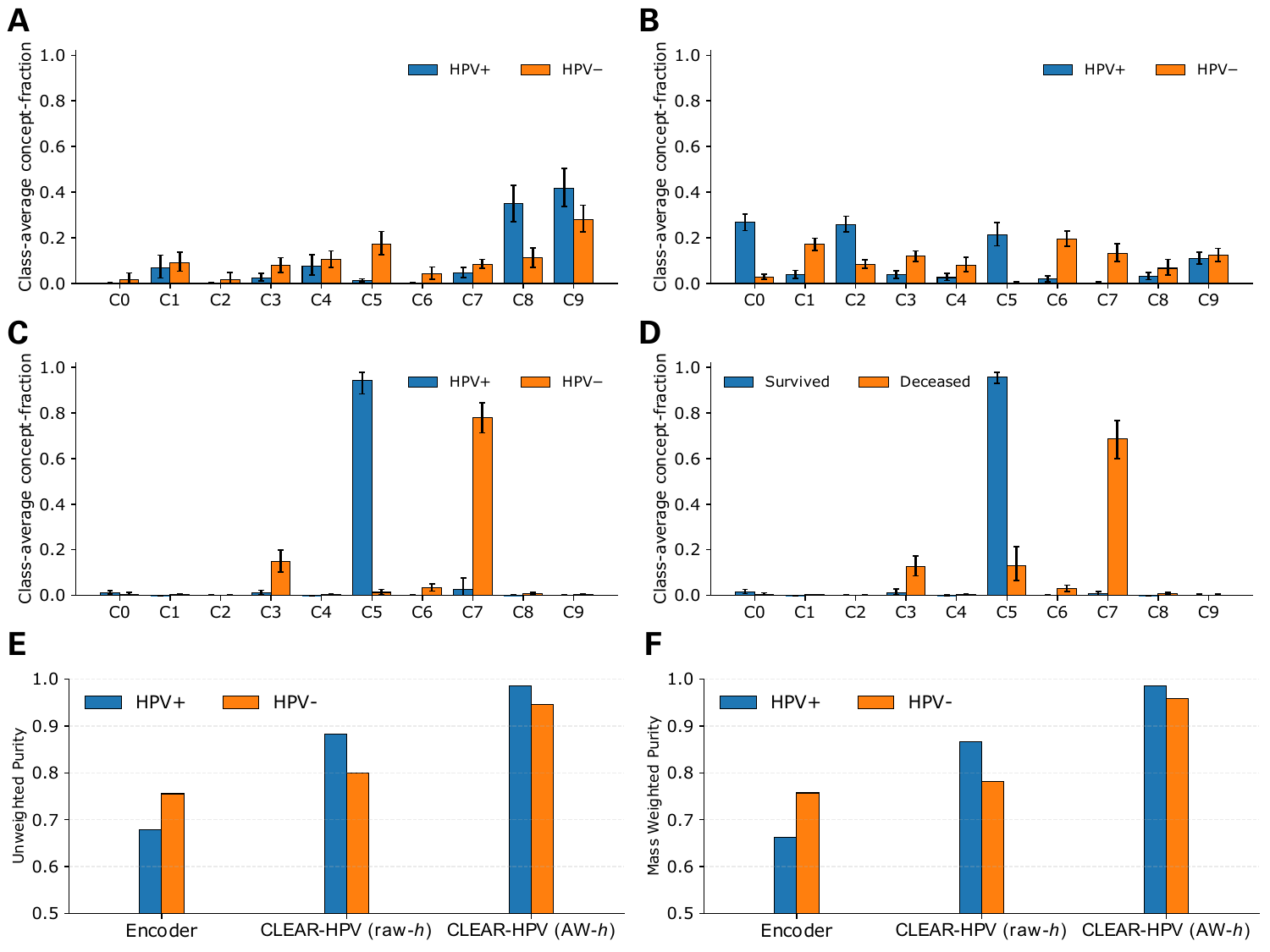}
    \caption{ {\textbf{Class-averaged concept-fraction vectors across concept-discovery settings on TCGA-HNSCC.} Concept-fraction vectors are computed per slide as the fraction of tiles assigned to each discovered concept, optionally weighted by MIL attention. These slide-level vectors are then averaged within each group to obtain class-averaged profiles that summarize cohort-level morphologic composition and highlight differences in concept prevalence across clinical groups. Panels (A--C) show group-averaged profiles for HPV-positive (blue) and HPV-negative (orange) cases, while panel (D) shows the corresponding averages for surviving (blue) and deceased (orange) cases.
    \textbf{(A)} Class-averaged concept-fraction vectors derived from encoder-feature clustering (non-$h$-space baseline).
    \textbf{(B)} Class-averaged concept-fraction vectors derived from CLEAR-HPV concepts in the MIL $h$-space using unweighted fractions, where all tiles contribute equally.
    \textbf{(C)} Class-averaged concept-fraction vectors derived from CLEAR-HPV concepts in the MIL $h$-space using attention-weighted fractions, where each tile contributes proportionally to its MIL attention score, yielding clearer separation between HPV-positive and HPV-negative cases.
    \textbf{(D)} Using the same CLEAR-HPV $h$-space concepts and attention-weighted fractions as in (C), class-averaged concept-fraction vectors are shown for surviving versus deceased cases.}
    \textbf{(E)} Unweighted purity (cluster-averaged class dominance) across methods, computed as the mean class-dominance ratio over clusters dominated by each class.
    \textbf{(F)} Mass weighted purity (dominant-cluster class-mass concentration) across methods, computed as the fraction of class-specific mass within clusters dominated by each class; higher values indicate stronger class coherence. Detailed definitions are provided in Supplemental Methods S1.
    }

  \label{fig:fractions}
\end{figure}

Baselines that do not operate in the $h$-space performed substantially worse. We find that heatmap-based grouping produced diffuse partitions with limited HPV separation, indicating that spatial saliency alone is insufficient for isolating meaningful morphologic patterns; the Dirichlet model emphasizes either global regularity or high flexibility, which can conflict with the localized, heterogeneous, and uneven distribution of HPV-associated morphologic patterns in histopathologic tissue, making these distributions difficult to fit reliably under weak slide-level supervision; encoder-based clustering achieved reasonable quantitative performance but showed weaker class-coherent organization than $h$-space-based concept discovery, both qualitatively and quantitatively. In particular, the added class-dominance analyses in \figref{fig:fractions}(E) and (F) show that encoder-space concepts exhibit substantially lower dominant-cluster purity and mass concentration than CLEAR-HPV concepts, indicating greater mixing of class-relevant morphology within clusters. Both metrics are defined in Supplemental Methods S1.

\paragraph{Recovery score: How well CLEAR-HPV retains predictive performance.} 
To further assess how well the discovered concepts retain the predictive performance of the backbone MIL model, we also compute a ``recovery score'' that quantifies how much predictive performance (e.g., accuracy and AUC) can be recovered using only the discovered concepts from different methods (including our CLEAR-HPV). (Details on how to compute the recovery score are included in the Methods section.) \figref{fig:compare} shows the results. CLEAR-HPV (AW-$h$) and CLEAR-HPV (raw-$h$) achieve the highest recovery scores, demonstrating that attention-weighted latent structure supports faithful, interpretable concept decompositions. Baselines such as heatmap-based grouping and Dirichlet clustering achieve much lower recovery scores, indicating a limited ability to preserve the original model's predictive performance. 

Together, these results show that concept discovery is most effective when performed directly in the attention-structured $h$-space, and that CLEAR-HPV can provide high-quality explanations of an MIL backbone's prediction by retaining its discriminative signal while providing compact, interpretable concept-based representations. These are strong results because they show that CLEAR-HPV (1) preserves the predictive performance of the interpreted model while reducing its high-dimensional features (e.g., 1536 dimensions) to only 10 interpretable concepts and (2) improves class-coherent organization of discovered concepts, as quantified by the our class-dominance metrics.

\subsection*{Interpretability}

While the preceding section evaluated how well different concept-discovery methods preserve predictive signal and performance using quantitative classification metrics, these metrics alone do not explain what morphologic patterns the models rely on. We therefore next examine the interpretability of the discovered concepts by analyzing how they are distributed across slides and how clinicians can use these concept-fraction vectors. This analysis shifts the focus from performance to biological meaning, asking whether the learned concepts correspond to coherent and clinically recognizable histopathologic patterns.

\paragraph{Class-averaged concept-fraction analysis.} To facilitate interpretability (explainability), we analyze concept-fraction vectors at the cohort level by averaging slide-level concept-fraction vectors within each group (e.g., HPV-positive and HPV-negative patients). These class-averaged concept-fraction vectors provide a compact quantitative summary of morphologic composition that can be directly compared across HPV status, survival outcomes, and cohorts. For example, an average concept-fraction vector over the HPV-positive patients provide insight on what concepts are most relevant to the positive predictions, and similarly for HPV-negative patients. A larger difference between these two vectors indicates that the discovered CLEAR-HPV concepts better distinguish HPV-positive from HPV-negative cases.

\begin{figure}[!t]
  \centering
  \includegraphics[width=\linewidth]{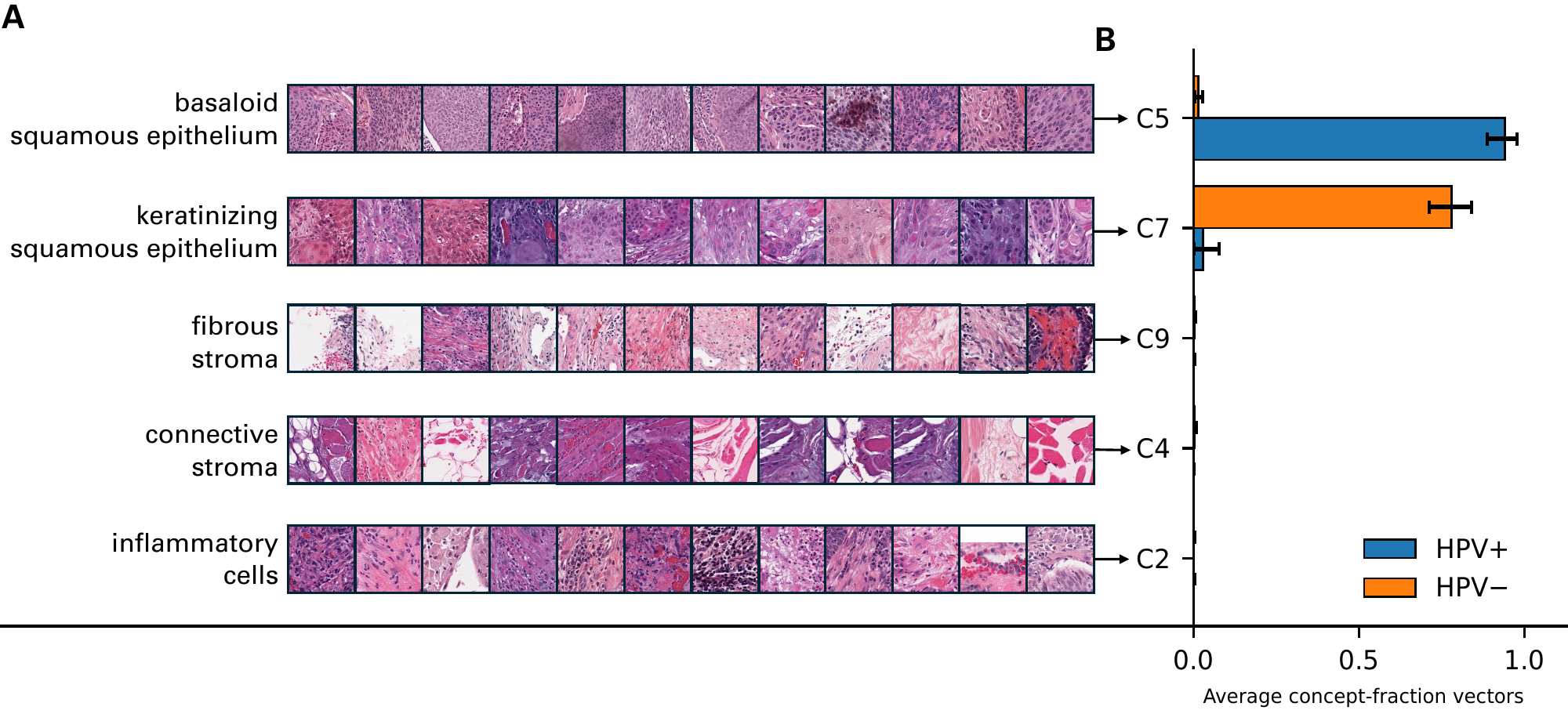}
\caption{
\textbf{Top tiles for key concepts discovered by CLEAR-HPV (A) and the corresponding slide-level distributions in the dataset TCGA-HNSCC (B).}
\textbf{(A)} Top (representative) tiles for five CLEAR-HPV concepts chosen for their consistent appearance and clear morphologic identity:
C5 (basaloid squamous epithelium),
C7 (keratinizing squamous epithelium),
C9 (fibrous stroma),
C4 (connective stroma), and
C2 (inflammatory cells).
\textbf{(B)} Average concept-fraction vectors for HPV-positive (\blue{blue}) and HPV-negative (\orange{orange}) slides, with 95\% bootstrap confidence intervals. 
The ``Basaloid'' Concept C5 is more prevalent in HPV-positive cases, while the ``keratinizing'' Concept C7 is more prevalent in HPV-negative cases. 
}
  \label{fig:topk}
\end{figure}

\figref{fig:fractions} shows the class-averaged concept-fraction vectors for HPV-positive and HPV-negative cases with $K=10$ concepts (C0--C9). Encoder-space concepts (\figref{fig:fractions}A) show little separation between classes. For CLEAR-HPV concepts discovered in the MIL $h$-space, unweighted concept fractions (\figref{fig:fractions}B; each tile contributes equally to its assigned concept) yield only modest class separation. In contrast, attention-weighted concept fractions (\figref{fig:fractions}C; each tile’s contribution to the slide-level fraction is weighted by its MIL attention score) produce a clearer dichotomy: HPV-positive slides show higher fractions of the basaloid/non-keratinizing concept (C5), whereas HPV-negative slides show higher fractions of the keratinizing concept (C7). Representative tiles supporting these morphologic interpretations are shown in \figref{fig:topk}, consistent with established histopathologic differences. Using the same attention-weighted fraction computation, \figref{fig:fractions}D shows a clear survival-associated discrepancy in concept composition. Survivors are enriched for C5, whereas deceased cases are enriched for C7. This pattern is consistent with the established prognostic advantage of HPV-driven tumors, even though the MIL backbone was trained only for HPV prediction.

To quantify class-coherent structure of discovered concepts, we introduce two metrics based on class-averaged concept-fraction vectors: unweighted purity, which measures the mean purity of class-dominant clusters, and mass weighted purity, which emphasizes whether class-specific signal is concentrated in high-mass clusters. Detailed definitions are provided in Supplemental Methods S1. Both metrics consistently show that attention-weighted CLEAR-HPV produces higher class-dominance and mass concentration than encoder-space concepts, indicating improved class-coherent organization of morphology.

\paragraph{Concept identity and representative morphology.} \figref{fig:topk} shows representative tiles for key concepts discovered by CLEAR-HPV (\figref{fig:topk}(A) and the corresponding slide-level distributions in TCGA-HNSCC (\figref{fig:topk}(B)). In \figref{fig:topk}(A), clinician review confirmed that these concepts correspond to inflammatory infiltrates (C2), benign stroma (C4), HPV-associated basaloid carcinoma (C5), HPV-negative keratinizing carcinoma (C7), and fibrous stroma (C9). These concepts appear consistently across datasets and form the basis for interpretable (explainable) slide-level composition profiles. In \figref{fig:topk}(B), we can see that the discovered Concept 5 (C5) and Concept 7 (C7) are highly relevant to HPV-positive and HPV-negative cases, respectively.   

\begin{figure}[!t]
  \centering
  \includegraphics[width=\linewidth]{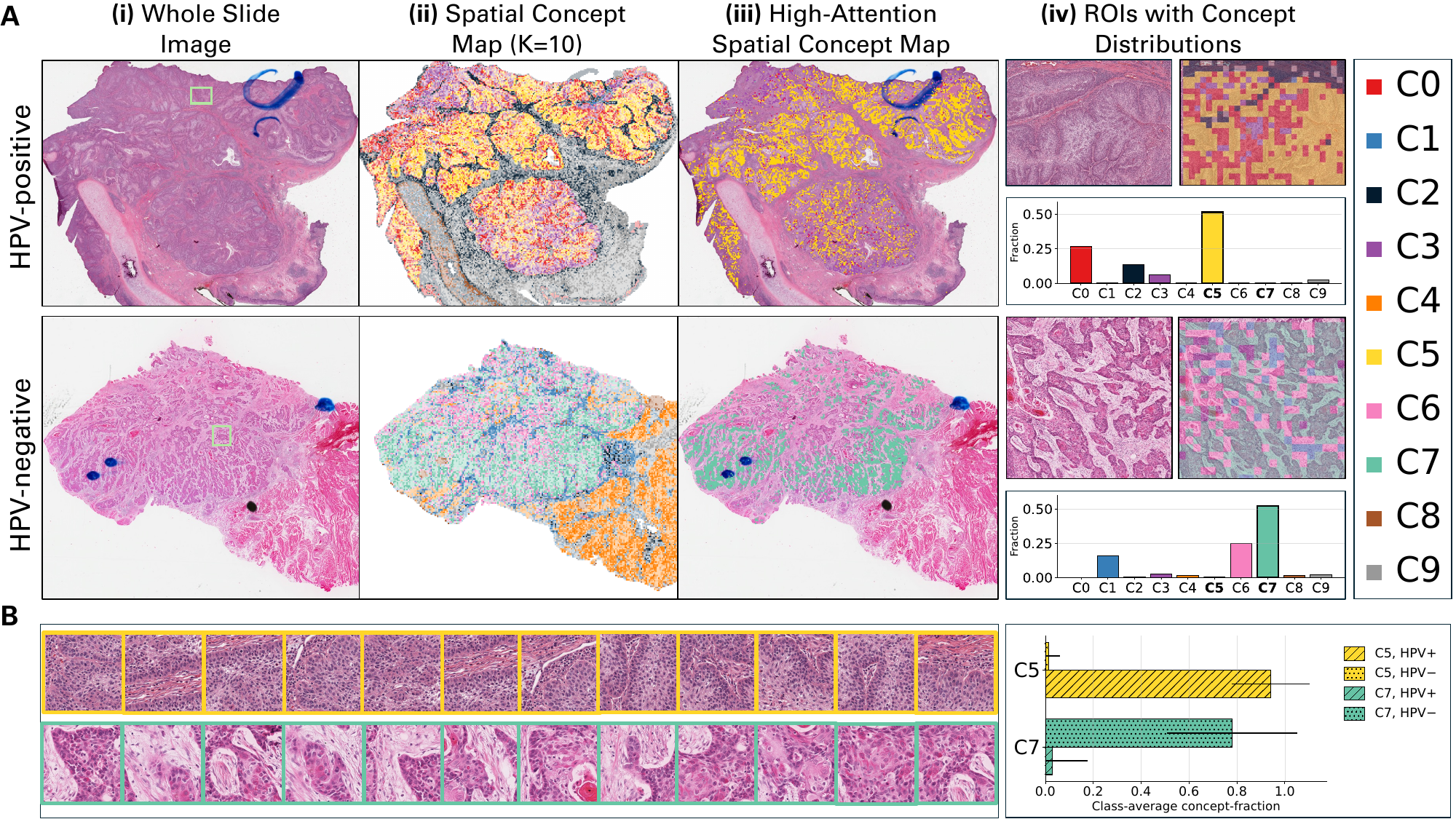}
    \caption{
        \textbf{Visualization of attention-weighted concept discovery using CLEAR-HPV.}
        \textbf{(A)} For representative HPV-positive and HPV-negative WSIs 
        from TCGA-HNSCC, we show, in \textbf{four columns}:
        (\textbf{\textit{i}}) the original H\&E whole slide image,
        (\textbf{\textit{ii}}) the $h$-space spatial concept map,
        (\textbf{\textit{iii}}) the high-attention spatial concept map, 
        and
        (\textbf{\textit{iv}}) regions of interest (ROIs) with their corresponding concept-fraction distributions produced by our CLEAR-HPV.
        \textbf{(B)} Representative tiles for two CLEAR-HPV concepts (C5 and C7), along with the average concept-fraction vectors for HPV-positive and HPV-negative slides of the entire dataset. We use different colors to represent different concepts consistently across all figures. Blue markings visible on the whole slide images in (A) are pre-existing annotation artifacts on physical slides by clinicians and can be omitted from interpretation.
    }
  \label{fig:clustering}
\end{figure}

\paragraph{Slide-level visualization of concepts discovered by CLEAR-HPV.} 
\figref{fig:clustering} visualizes concepts within two representative HPV-positive and HPV-negative slides. High-attention maps (Column 3 of~\figref{fig:clustering}(A)) are obtained by ranking tile-level attention scores and retaining only top-scoring tiles, capturing regions most emphasized by the MIL backbone. Within these areas, HPV-positive slides consistently show the ``basaloid'' concept C5 (in yellow), while HPV-negative slides show the ``keratinizing'' concept C7. Background concepts such as fibrous or benign stromal tissue (e.g., C4 and C9), as indicated by their representative tiles in \figref{fig:topk}, also appear consistently across slides and provide visual context for surrounding tumor regions. These representative tiles are selected by their distance to concept center (see details in the Methods section).

To quantify concept composition in highlighted, key areas of a slide, we define regions of interest (ROIs) as the contiguous high-attention regions outlined by the green box in \figref{fig:clustering}A(i), with a zoomed view shown in \figref{fig:clustering}A(iv) The concept-fraction distributions in the regions of interest (ROIs), shown in the green box in \figref{fig:clustering}(A)(i) and in \figref{fig:clustering}(A)(iv), match the dataset-level average concept-fraction vectors in \figref{fig:clustering}(B)(right): C5 is enriched in HPV-positive cases, and C7 is enriched in HPV-negative cases. Together, these results show that (1) our CLEAR-HPV can provide slide-level concept explanations for HPV prediction while also highlighting important regions in the slide, and that (2) our CLEAR-HPV produces consistent concepts at the slide and dataset levels; slide-level average concept-fraction vectors within high-attention regions align with their dataset-level (global) average concept-fraction vectors across the cohort.

\subsection*{Comparison of concept discovery across MIL backbones}
Our CLEAR-HPV framework is compatible with arbitrary attention-based MIL neural network architectures. In addition to CLAM, we evaluated three MIL backbones, including two widely used MIL backbones, i.e., attention-based MIL (ABMIL)\cite{ITW:2018} and Transformer-based MIL (TransMIL) \cite{shao2021transmil}, as well as a multi-head attention-based MIL model (MHMIL), to investigate how backbone choices influence the structure of the latent $h$-space and the resulting CLEAR-HPV concepts. 

ABMIL employs a gated attention mechanism to learn instance-level importance weights and aggregates tile embeddings via a weighted sum, providing a simple and effective attention-based pooling strategy. TransMIL extends the MIL paradigm by modeling global contextual relationships among instances using Transformer layers, enabling long-range dependency modeling across tiles within a WSI. MHMIL is a multi-head attention-based MIL model inspired by CLAM; it replaces the single attention head with multiple parallel heads, allowing different heads to capture complementary tissue patterns while preserving the underlying MIL aggregation structure. By applying CLEAR-HPV to ABMIL, TransMIL, and MHMIL, we assess its robustness of concept discovery across distinct architectural inductive biases.

Using the same evaluation procedure as in the CLAM-based experiments, each slide is represented by a concept-fraction vector and evaluated with a concept-fraction classifier across all backbones. As shown in Table~\ref{tab:backbone_main}, CLEAR-HPV consistently preserves slide-level predictive performance across multiple attention-based MIL architectures, even though it replaces the original \emph{high-dimensional} latent representations (i.e., $1536$ dimensions), which are \emph{not} interpretable at the concept level, with substantially more \emph{compact} (i.e., $10$ dimensions) and \emph{interpretable} concept representations. These are therefore \emph{strong} results. Across ABMIL, TransMIL, and MHMIL, CLEAR-HPV achieves AUC ($\approx$ 0.84--0.88) and accuracy ($\approx$ 0.76--0.81) that closely match those of the corresponding original backbone models (AUC $\approx$ 0.88--0.90, ACC $\approx$ 0.78--0.82). Importantly, we observe the following:
\begin{itemize}
\item Performance retention is consistent across architectures with distinct attention mechanisms and modeling complexity, indicating that the discovered concepts capture backbone-agnostic, diagnostically relevant morphology rather than architecture-specific artifacts. 
\item High performance is achieved when compressing the $1536$-dimensional representation from the MIL backbone to only $10$ interpretable concept dimensions, demonstrating that CLEAR-HPV retains discriminative signal in a substantially more compact and interpretable representation, without reliance on high-dimensional, uninterpretable embeddings.
\end{itemize}

\paragraph{Encoder dependence.} 
In addition to backbone robustness, we evaluated whether CLEAR-HPV depends on the choice of feature encoder by replacing the UNI encoder with a conventional ResNet50. CLEAR-HPV maintains comparable performance under this change and continues to produce coherent concept representations, indicating that the proposed framework is not tied to a specific foundation model encoder. Detailed results are provided in Supplementary Table S5.

\begin{table}[!t]
\centering
\footnotesize
\setlength{\tabcolsep}{4pt}
\renewcommand{\arraystretch}{1.4}

\begin{tabular}{llccccccc}
\toprule
\textbf{Backbone} & \textbf{Method} & \textbf{Feat. Dim.} & \textbf{Interpret.}
& \textbf{ACC} & \textbf{AUC} & \textbf{Prec} & \textbf{Rec} & \textbf{F1} \\
\midrule
\multicolumn{7}{l}{\textit{Original backbone performance (no concept discovery)}} \\
\hline
ABMIL & original & 1536 & \xmark
 & 0.818\pmv{0.051}
 & 0.879\pmv{0.062}
 & 0.867\pmv{0.098}
 & 0.725\pmv{0.104}
 & 0.765\pmv{0.068} \\

TransMIL & original & 1536 & \xmark
 & 0.784\pmv{0.057}
 & 0.899\pmv{0.052}
 & 0.800\pmv{0.098}
 & 0.698\pmv{0.085}
 & 0.734\pmv{0.072} \\

 MHMIL & original & 1536 & \xmark
 & 0.791\pmv{0.079}
 & 0.879\pmv{0.068}
 & 0.825\pmv{0.134}
 & 0.685\pmv{0.160}
 & 0.723\pmv{0.116} \\

\midrule
\multicolumn{7}{l}{\textit{CLEAR-HPV concept discovery}} \\
\hline

ABMIL 
& AW-$h$ & 10 & \cmark
& 0.799\pmv{0.087}
& 0.859\pmv{0.081}
& 0.769\pmv{0.115}
& 0.767\pmv{0.222}
& 0.751\pmv{0.133}\\

ABMIL & raw-$h$ & 10 & \cmark
 & 0.810\pmv{0.068}
 & 0.868\pmv{0.086}
 & 0.811\pmv{0.145}
 & 0.763\pmv{0.169}
 & 0.769\pmv{0.102 }\\

TransMIL & AW-$h$ & 10 & \cmark
 & 0.765\pmv{0.077}
 & 0.863\pmv{0.089}
 & 0.738\pmv{0.071}
 & 0.703\pmv{0.179}
 & 0.713\pmv{0.111} \\

TransMIL & raw-$h$ & 10 & \cmark
 & 0.784\pmv{0.089}
 & 0.841\pmv{0.084}
 & 0.803\pmv{0.149}
 & 0.678\pmv{0.147}
 & 0.727\pmv{0.118} \\

MHMIL & AW-$h$ & 10 & \cmark
& 0.782\pmv{0.145}
& 0.859\pmv{0.144}
& 0.813\pmv{0.229}
& 0.669\pmv{0.242}
& 0.716\pmv{0.204} \\
 
MHMIL & raw-$h$ & 10 & \cmark
& 0.794\pmv{0.100}
& 0.879\pmv{0.129}
& 0.859\pmv{0.140}
& 0.669\pmv{0.273}
& 0.713\pmv{0.178} \\
\hline
\end{tabular}

\caption{
{\textbf{Original backbone performance vs. CLEAR-HPV concept discovery across MIL backbones (on TCGA-HNSCC).} For each attention-based MIL backbone, we report the original backbone performance and the corresponding results obtained by applying CLEAR-HPV for concept discovery on the latent $h$-space. \textit{Feat.\ Dim.} denotes the dimensionality of the slide-level representation used for classification, and \textit{Interpret.} indicates whether the representation yields explicit, human-interpretable morphologic concepts. Across all backbones, CLEAR-HPV achieves strong performance comparable to the original models. For MHMIL, performance metrics for individual attention heads and alternative head-aggregation strategies are provided in in Table S4.} 
} 

\label{tab:backbone_main}
\end{table}

\subsection*{Cross-cohort generalization and robustness}
CLEAR-HPV concepts transfer reliably across datasets that differ in anatomic sites, staining profiles, and class distributions. All cross-cohort experiments were conducted in a zero-shot setting: concept clusters, attention weights, and the concept-fraction classifier were trained exclusively on TCGA-HNSCC and were not adapted, recalibrated, or tuned using any slides from TCGA-CESC or CPTAC-HNSCC. This design ensures that evaluations on the external cohorts measure genuine generalization rather than domain-specific adjustment. These results should be interpreted in terms of signal preservation and interpretability, rather than improvement in predictive performance.

\begin{figure}[!t]
  \centering
  \includegraphics[width=\linewidth]{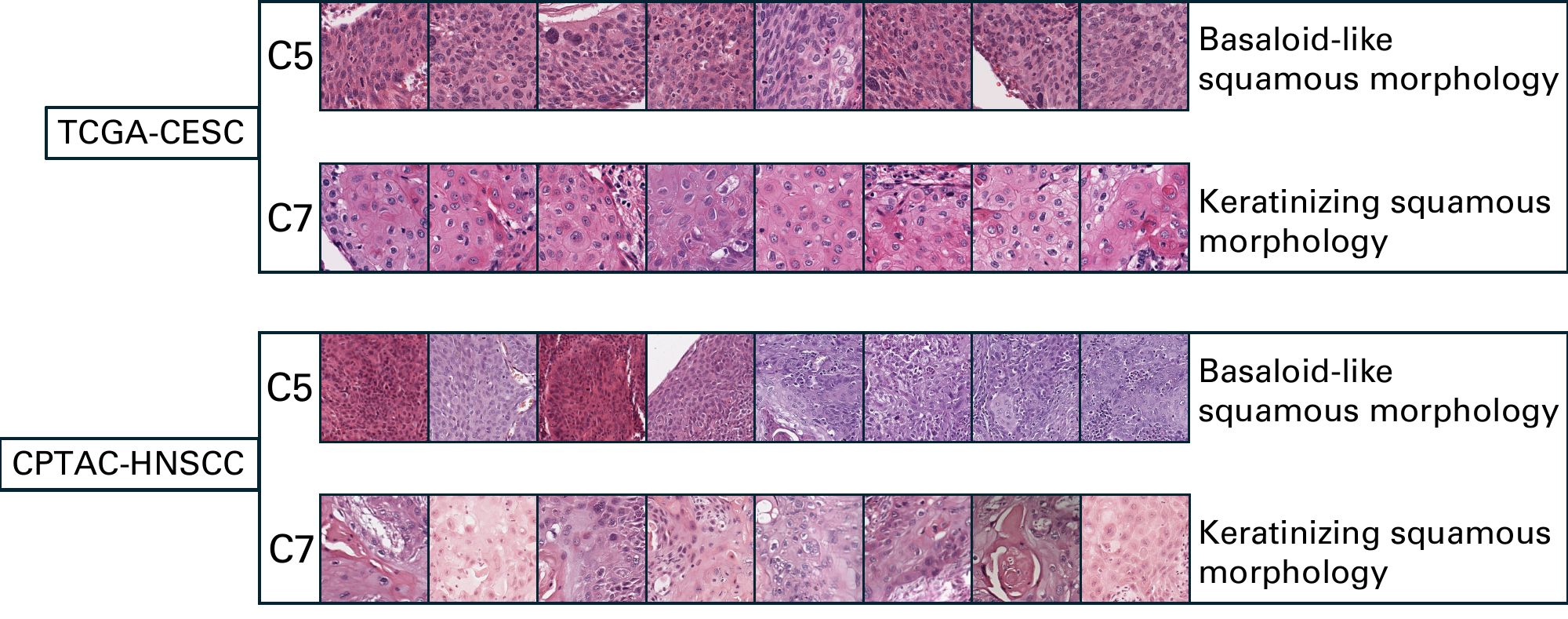}
  \caption{
    \textbf{Cross-cohort consistency of HPV-related concepts among top-$8$ tiles.} 
    We show representative high-attention tiles for the HPV-positive-related ``basaloid'' concept C5 and the ``keratinizing'' concept C7 from two external cohorts, TCGA-CESC (\textbf{top}) and CPTAC-HNSCC (\textbf{bottom}). Across both datasets, C5 consistently reflects basaloid morphology characteristic of HPV-positive tumors, while C7 reflects keratinizing morphology typical of HPV-negative tumors. The consistency of these signatures across independent cohorts demonstrates that CLEAR-HPV identifies stable, biologically meaningful concepts that generalize beyond the training dataset. 
}
  \label{fig:excohort}
\end{figure}

\paragraph{Transferring to the external cohort TCGA-CESC.} \tabref{tab:hnscctrain_cesc_test_cluster_small} shows the results when models trained on TCGA-HNSCC were applied to the external cohort TCGA-CESC. Both CLEAR-HPV variants, AW-$h$ and raw-$h$, retain substantial predictive signal under this substantial biological shift in tissue site. Note that TCGA-CESC differs substantially from TCGA-HNSCC in both anatomy and cohort composition: it is dominated by HPV-positive tumors and exhibits histomorphologic patterns distinct from HNSCC \cite{TCGA_CESC_2017}. Under this extreme imbalance, the CLAM baseline achieved only modest performance (ACC $\approx$ 0.59, AUC $\approx$ 0.68). Due to the strong class imbalance in TCGA-CESC, threshold-based metrics such as F1 and Recall should be interpreted with caution, as they can be influenced by majority-class bias.
Interestingly, both CLEAR-HPV variants, with their discovered interpretable concepts and concept-fraction vectors, maintain comparable AUC while achieving higher accuracy (ACC $\approx$ 0.75), and, more importantly, preserved CLAM’s extremely high precision (Prec $\approx$ 0.96); they also achieved higher sensitivity and F1 than the CLAM base model. The attention-weighted (AW) variant achieved the highest F1 ($\approx$ 0.86) and Sensitivity ($\approx$ 0.78), while the raw-$h$-space variant obtained the highest AUC ($\approx$ 0.67) among CLEAR-HPV variants. However, AUC indicates a modest degradation in ranking performance under domain shift, consistent with our interpretation that CLEAR-HPV preserves interpretable structure rather than improving global discrimination. These results show that CLEAR-HPV captures HPV-related morphologic signals that remain predictive, while maintaining interpretability under major shifts in tissue architecture. Together with the cross-cohort concept consistency shown in Fig.~\ref{fig:excohort}, these results indicate that CLEAR-HPV preserves transferable morphologic structure even when ranking performance degrades modestly.

\paragraph{Transferring to the external cohort CPTAC-HNSCC.} Similarly, \tabref{tab:cptac_singleclass_small} shows the results when models trained on TCGA-HNSCC were applied to the external cohort CPTAC-HNSCC, demonstrating CLEAR-HPV's robustness to technical and institutional variability. Because this cohort contains only HPV-negative tumors, accuracy (ACC) is the only relevant metric. Both CLEAR-HPV variants outperformed CLAM (ACC $\approx$ 0.82--0.90), despite pronounced differences in staining, slide preparation, and scanner characteristics \cite{edwards2015cptac} between these two cohorts. This improvement indicates that our CLEAR-HPV's concept-fraction representations (vectors) remain stable under technical domain shift.

\begin{table}[!t]
\centering
\footnotesize
\setlength{\tabcolsep}{11pt}
\renewcommand{\arraystretch}{1.15}

\begin{tabular}{lccccc}
\hline
\textbf{Method} & \textbf{ACC} & \textbf{AUC} & \textbf{Prec} & \textbf{Rec} & \textbf{F1} \\
\hline

CLAM
 & 0.593\pmv0.083
 & 0.684\pmv0.154
 & 0.977\pmv0.029
 & 0.581\pmv0.176
 & 0.716\pmv0.073 \\

CLEAR-HPV (AW-$h$)
 & 0.756\pmv0.069
 & 0.650\pmv0.067
 & 0.958\pmv0.013
 & 0.777\pmv0.083
 & 0.855\pmv0.049 \\

CLEAR-HPV (raw-$h$)
 & 0.748\pmv0.056
 & 0.671\pmv0.061
 & 0.960\pmv0.012
 & 0.765\pmv0.066
 & 0.850\pmv0.041 \\
\hline
\end{tabular}

    \caption{
    \textbf{Cross-cohort generalization from TCGA-HNSCC to TCGA-CESC.}
    Under severe domain shift and strong HPV-positive class imbalance, CLEAR-HPV retains substantial predictive signal in an interpretable concept space, although AUC indicates modest degradation in ranking performance relative to CLAM. raw-$h$ achieves the highest AUC among the CLEAR-HPV variants, while AW-$h$ provides the highest F1 and Recall. Results for additional metrics are provided in Table S6 
    }
\label{tab:hnscctrain_cesc_test_cluster_small}
\end{table}

\begin{table}[!t]
\centering
\footnotesize
\setlength{\tabcolsep}{27pt}   % widen columns
\renewcommand{\arraystretch}{1.30}
    \begin{tabular}{lccc}
    \hline
    \textbf{Metric} 
    & \textbf{CLAM} 
    & \textbf{CLEAR-HPV (AW-$h$)} 
    & \textbf{CLEAR-HPV (raw-$h$)} \\
    \hline
    ACC 
    & 0.704\pmv0.120 
    & 0.824\pmv0.146 
    & 0.896\pmv0.094 \\
    \hline
    \end{tabular}

\caption{
\textbf{External validation on CPTAC-HNSCC.}
Only accuracy is reported because the cohort contains only HPV-negative cases.
CLEAR-HPV preserves the predictive information of the CLAM base model when transferred to an external cohort.
}
\label{tab:cptac_singleclass_small}
\end{table}

\paragraph{Morphologic consistency of concepts across cohorts.} \figref{fig:excohort} illustrates that the HPV-associated ``basaloid'' concept (C5) and the ``keratinizing'' concept (C7) retain their characteristic morphology in high-attention regions across both external cohorts, TCGA-CESC and CPTAC-HNSCC. The same concepts, basaloid tiles for C5 and keratinizing squamous morphology for C7, appear consistently even in these external cohorts not seen during training. These cross-cohort results demonstrate that CLEAR-HPV captures stable, biologically meaningful morphologic structures that generalize across different squamous tumor types and institutional settings.

\subsection*{Survival prediction based on CLEAR-HPV concepts}
HPV status is a well-established prognostic marker in squamous cell carcinoma, with HPV-positive tumors exhibiting significantly better survival outcomes \cite{Mobadersany2018PNAS}. We therefore assessed whether CLEAR-HPV concepts retain this prognostic signal by evaluating survival prediction in TCGA-HNSCC using only the concept-fraction vectors derived using CLEAR-HPV applied to an HPV-trained backbone. No retraining or survival-specific adaptation was performed; therefore the survival model receives solely the morphologic composition learned from HPV supervision. 

As shown in \tabref{tab:survival_clam_vs_kmeans}, {the attention-weighted concept vectors produced by CLEAR-HPV achieve competitive, though slightly lower, predictive performance compared to the base model CLAM. While the explained model, CLAM, achieves the highest AUC ($\approx$ 0.840), both CLEAR-HPV variants (which explain CLAM's predictions) retain broadly comparable discriminative ability despite operating in a substantially lower-dimensional space.} Note that this is already strong results because the CLEAR-HPV's concept-fraction vectors have only $10$ dimensions; they are both compact and interpretable. In contrast, the original CLAM model uses embeddings with {$1536$} dimensions, and are not interpretable.

These results indicate that CLEAR-HPV concepts capture broader morphologic structure that remains informative for patient outcome, even though the concepts were originally learned for HPV prediction. The concept-fraction vectors provide a compact and interpretable summary of tumor and stromal composition, allowing links between morphologic patterns and adverse prognosis to be examined. Together, these findings show that reorganizing the MIL latent space into CLEAR-HPV concepts preserves survival-relevant information, even when supervision is based entirely on HPV labels.

\begin{table}[!t]
\centering
\footnotesize
\setlength{\tabcolsep}{8pt}
\renewcommand{\arraystretch}{1.15}

\begin{tabular}{lccccc}
\hline
\textbf{Method} & \textbf{ACC} & \textbf{AUC} & \textbf{Prec} & \textbf{Rec} & \textbf{F1} \\
\hline

CLAM
 & 0.734\pmv0.086
 & 0.840\pmv0.081
 & 0.668\pmv0.160
 & 0.673\pmv0.188
 & 0.628\pmv0.143 \\
 
CLEAR-HPV (AW-$h$)
 & 0.715\pmv0.121
 & 0.785\pmv0.134
 & 0.665\pmv0.209
 & 0.558\pmv0.147
 & 0.594\pmv0.168 \\

CLEAR-HPV (raw-$h$)
 & 0.700\pmv0.139
 & 0.766\pmv0.114
 & 0.675\pmv0.229
 & 0.530\pmv0.198
 & 0.563\pmv0.192 \\
\hline
\end{tabular}

\caption{
\textbf{Survival prediction on TCGA-HNSCC.}
Survival prediction using CLEAR-HPV concept representations (i.e., concept-fraction vectors) is compared with the base model CLAM. Results show that CLEAR-HPV's concept-fraction vectors preserve the prognostic information contained in the
original MIL embeddings of CLAM. {Results for more metrics are provided in Table S7.}
}
\label{tab:survival_clam_vs_kmeans}

\end{table}

\section*{DISCUSSION}

This study demonstrates that the latent space learned by attention-based MIL models contains biologically structured information that becomes interpretable when it is reorganized through our proposed CLEAR-HPV framework. The latent $h$-space encodes tile-level morphology before slide-level aggregation. Weighting these embeddings by learned attention scores highlights the regions most relevant to the prediction and enables CLEAR-HPV to recover consistent and generalizable morphologic concepts without tile-level supervision or human annotation. CLEAR-HPV is not designed to improve classification accuracy, but to provide a structured and interpretable view of the latent space learned by the model.

Our findings further demonstrate that CLEAR-HPV is not tied to a specific attention mechanism or MIL architecture, but can be applied consistently across diverse attention-based backbones. Despite differences in model architecture and attention formulation, CLEAR-HPV relies on the attention learned for the prediction task to organize the latent $h$-space, resulting in stable and robust morphologic concept discovery. This architectural robustness indicates that CLEAR-HPV functions as a general concept discovery framework that can be integrated with a wide range of attention-based MIL models without requiring model-specific modification.

CLEAR-HPV concepts demonstrate cross-cohort consistency across TCGA-HNSCC, TCGA-CESC, and CPTAC-HNSCC in a strict zero-shot transfer setting. Despite substantial variation in staining, scanner characteristics, and cohort composition~\cite{Stacke2021DomainShiftHistopathology, Zarella2019PracticalWSI}, the same ``basaloid'' and ``keratinizing'' concepts appeared consistently across cohorts (datasets). This reproducibility and stability indicate that CLEAR-HPV captures transferable morphologic signals associated with HPV status rather than dataset-specific artifacts, even under substantial domain shift, although some degradation in ranking performance is observed in certain transfer settings. Importantly, these results reinforce that CLEAR-HPV is designed to preserve predictive signal in an interpretable representation rather than to optimize classification performance, and thus minor changes in standard metrics should be interpreted in the context of this trade-off.

The concept-fraction vectors produced by CLEAR-HPV also retained prognostic information (important for survival analysis) in TCGA-HNSCC, even though the concepts were learned solely from HPV supervision without any concept-level annotation or supervision. This shows that reorganizing the MIL latent space can preserve broader tumor biology and that interpretable concept profiles can serve as compact descriptors for downstream clinical tasks.

While recent progress in pathology foundation models (e.g., contrastive histology encoders~\cite{SelfSupervisedHisto}, TCv2~\cite{nicke2025tissueconceptsv2supervised}, UNI~\cite{chen2024uni}, CONCH~\cite{Lu2024NatMed}, and GigaPath~\cite{xu2024gigapath}) provides powerful feature extractors trained at scale, our results highlight that expressive embeddings alone do not guarantee interpretability (or explainability). Direct clustering of foundation-model features produced reasonable quantitative performance but did not reveal coherent morphologic structure. The attention-weighted $h$-space, by contrast, combines the representational richness of these encoders with task-specific focus, enabling concept discovery that aligns with the regions most relevant to the prediction. This intermediate representation offers a practical pathway toward interpretable foundation-model pipelines.

While CLEAR-HPV organizes latent representations into structured and quantitatively coherent concepts, assigning semantic labels to these concepts currently requires expert interpretation. Future work may explore the integration of large language models (LLMs) or multimodal vision-language models to assist in automatic concept description or naming, for example by summarizing representative tiles associated with each concept. Such approaches could improve scalability and usability, but any automatically generated annotations would still require expert validation for clinical use.

In summary, CLEAR-HPV provides an interpretable link between slide-level predictions and human-understandable histology by restructuring the attention-mediated latent space into discrete morphologic concepts. These concepts are biologically coherent, reproducible across cohorts, and informative for classification and prognosis. More broadly, our findings show that the attention-structured latent space can support concept-level interpretability in whole-slide imaging across diverse MIL backbone architectures, offering a general framework for interpretable MIL. We expect that similar strategies will extend to additional molecular and prognostic tasks and will contribute to interpretable representation learning in computational pathology.

\subsection*{Limitations of the study}

This study has several limitations. First, the quality of the recovered concepts is influenced by the choice of MIL backbones; models with diffuse and weaker attention may offer slightly weaker structure for unsupervised separation of different concepts. Second, clustering is performed in a latent space without tile-level labels; therefore certain distinctions may remain subtle and benefit from expert interpretation. In particular, clinically actionable interpretation of the discovered concepts requires expert pathology review. Third, our pipeline relies on pretrained encoders and fixed attention maps, following the standard histopathology MIL paradigm (e.g., CLAM), where tile encoders are not jointly optimized during slide-level training. A unified end-to-end optimization of the encoder, attention mechanism, and concept structure may yield more refined and semantically consistent concepts, and remains an important direction for future work. Future work will incorporate spatial context, richer supervisory signals, and multimodal information to expand the biological insight provided by the CLEAR-HPV framework.

\newpage

%%%  The methods (also/formerly called "experimental 
%%%  procedures") should appear 
%%%  immediately after the discussion. Subheadings 
%%%  may be customized. Please consult your handling 
%%%  editor if you have questions about what content 
%%%  should appear here. 

\section*{METHODS}

% \hao{do not leave the space between two section heads empty. we need to mention Figure 1 and briefly introduce our overall framework. I have written something below.}

\figref{fig:main} shows the overview of our CLEAR-HPV framework. Whole slide images (WSIs) are first decomposed into fixed-size tiles, encoded with a pretrained ViT or CNN, and converted into patch-level feature embeddings (\figref{fig:main}(A)). 
An attention-based MIL classifier then projects embeddings into an $h$-space latent representation and uses multi-head attention to compute tile-level contributions, according to which tile-level embeddings are then pooled into a single slide-level embedding for HPV prediction (\figref{fig:main}(B)). 

Given an arbitrary attention-based MIL backbone model above, our CLEAR-HPV then performs annotation-free concept discovery on attention-weighted $h$-space embeddings to identify coherent morphologic concepts (\figref{fig:main}(C)). 
Using the discovered concepts, CLEAR-HPV can:
\begin{itemize}
\item compute the class-averaged concept-fractions to quantify the relative abundance of each concept for HPV-positive and HPV-negative slides within a given cohort; using attention weights can produce concepts that better distinguish between HPV-positive and HPV-negative cases (\figref{fig:main}(D)),
\item generate representative tiles that illustrate the characteristic morphology captured by each discovered concept (\figref{fig:main}(E)), 
\item generate spatial concept maps to visualize the distribution of concepts across the WSIs, revealing their spatial organization (\figref{fig:main}(F)), and
\item obtain slide-level concept-fraction vectors to provide an interpretable representation that supports a concept-fraction classifier, which recovers MIL predictive performance while offering concept-level explanations (\figref{fig:main}(G)).
\end{itemize}

% Below we describe different modules in detail.

Below, we provide details on each aformentioned module, evaluation, and implementation. 

\subsection*{Cohort curation and pathology criteria}

TCGA-HNSCC WSIs were curated by removing slides with poor staining, 
marked artifacts, or inadequate focus, using a combination of metadata filtering and manual quality inspection. For TCGA-CESC, we followed the pathology quality-control criteria described in TCGA\cite{TCGA_CESC_2017} and retained only diagnostic slides with adequate tumor content and acceptable image quality. CPTAC-HNSCC slides were included as provided, since this cohort had already underwent standardized acquisition and quality review through the CPTAC pipeline.

\subsection*{Data and preprocessing}
We used diagnostic WSIs from TCGA-HNSCC, TCGA-CESC, and CPTAC-HNSCC. Tissue regions were identified 
using standard color-based masking, and each WSI was tiled into non-overlapping patches (256$\times$256) at 
the full-resolution layer (level 0), as shown in~\figref{fig:main}(A). For each tile, we extracted a fixed embedding 
\(\mathbf{x}_i \in \mathbb{R}^{D_{\text{in}}}\) using the pretrained UNI encoder 
\cite{chen2024uni} (\(D_{\text{in}} = 1536\)) without any color normalization. Each slide with $N$ tiles is represented as an 
\(N \times D_{\text{in}} \) matrix of tile features. We have run experiments with color normalization too, but it turned out that color normalization does not make a difference, probably because the UNI encoder is already robust to different color schemes. Therefore we chose to run all experiments without any color normalization.

\subsection*{MIL backbones and the $h$-space}

\paragraph{Different MIL backbones.}We evaluated four attention-based MIL architectures, including {CLAM, ABMIL, TransMIL, and MHMIL}. All models share a unified projection layer that maps encoder features into a common latent $h$-space with the same dimensionality, 
enabling direct and controlled comparison of the resulting representations across backbones, 
% enabling direct comparison of latent space and attention structure across architectures. 
See~\figref{fig:main}(B) for a typical example of attention-based MIL model.

\paragraph{Attention-structured $h$-space.} 
As shown in~\figref{fig:main}(B), we define the $h$-space (the blue box) as the intermediate embedding space produced by the MIL
backbone prior to slide-level aggregation. For each tile \(i\), the backbone maps the encoder feature \(\mathbf{x}_i\) to an embedding \(\mathbf{h}_i = f_\theta(\mathbf{x}_i)\) and produces a raw attention score \(a_i\) (i.e., the logit before ``softmax''). 
% \hao{This is confusing: is $a_i$ already after softmax? if so, you do not need another softmax in the equation below. If not, you cannot call it "attention", you can maybe say  "a raw attention score (i.e., the logits before ``softmax'')"}
% \hao{for inline equations please use $XX$}
{Attention weights are obtained via slide-wise softmax over the $N$ tiles in the slide, followed by a rescaling step:}

% \[
\begin{align*}
\tilde{\alpha}_i = \frac{\exp(a_i)}{\sum_{j=1}^N \exp(a_j)}, \qquad
\alpha_i = N \cdot \tilde{\alpha}_i \qquad
% \frac{1}{N}\sum_{i=1}^N \alpha_i = 1.
\end{align*}
% \]

% \hao{please mention how you choose N, i.e., what value do you set it to. Is $N$ the number of tiles for each slide, so $N$ is different for each slide? If so, clarify this. Also the third equation is irrelevant, you can remove it, right?}

% \hao{for full equations please use the "align" environment (see my modification above}

{$\tilde{\alpha}_i$ denotes the normalized attention weight, and $N$ denotes the number of tiles extracted from the current slide. The rescaling sets the \emph{mean} weight within each slide to $1$ (since $\frac{1}{N}\sum_{i=1}^N \alpha_i = 1$), which makes the \emph{average} magnitude of attention weights comparable across slides with different numbers of tiles while preserving the relative ranking of tiles within each slide.} The \textbf{resulting attention-structured $h$-space representation of a slide} is given by

\begin{align*}
H = \{(\mathbf{h}_i, \alpha_i)\}_{i=1}^N
\end{align*}

% \hao{This is contradictory. According to the equation above, N is the number of tiles in each slide. So the weight is NOT invariant to the number of tiles?}

This standardized $h$-space representation above serves as the input to CLEAR-HPV’s unsupervised
concept discovery procedure, which analyzes the attention-weighted distribution
of tile embeddings to reveal structured morphologic patterns learned by the MIL
model.

\subsection*{Concept discovery module}

% \sout{CLEAR-HPV performs \emph{unsupervised} concept discovery by partitioning the attention-structured $h$-space into a
% fixed set of $K$ latent morphologic concepts, \emph{without needing any concept annotation}. To capture complementary aspects of structure in
% the tile distribution, we employ four unsupervised methods of increasing modeling capacity:
% (1) hard Euclidean assignment, 
% (2) attention-weighted extensions, 
% (3) optimal-transport regularization, and 
% (4) a probabilistic Dirichlet concept model with amortized inference. }
% \hao{Do these four correspond to the four methods in Table 2? I guess not? (1) you need to describe each method in Table 2 (2) for each method in this section you need to have experimental results (3) the method names in Table 2 and this section have to match}
% All methods operate on the tile-level embeddings $\{(\mathbf{h}_i, \alpha_i)\}$ and are trained in a streaming fashion to support whole-slide scale.

% In this section, we provide details on our CLEAR-HPV's concept discovery module and baseline methods evaluated in this manuscript. 

CLEAR-HPV performs unsupervised concept discovery by clustering the attention-structured $h$-space representation into a fixed set of $K$ latent morphologic concepts, \emph{without requiring any concept-level annotations (labels)}. In Table 2, we compare CLEAR-HPV (AW-$h$) and CLEAR-HPV (raw-$h$) against several unsupervised baselines, including Heatmap, Dirichlet Concepts, and Encoder Concepts. We describe each method in detail below. All methods operate on tile-level $h$-space representation $\{(\mathbf{h}_i, \alpha_i)\}$ and are designed to scale to whole-slide inference.

\paragraph{CLEAR-HPV (raw-$h$).}
The raw variant of CLEAR-HPV, i.e., \emph{CLEAR-HPV (raw-$h$)}, discovers $K$ morphologic concepts by clustering tile embeddings in the $h$-space using a scalable
two-stage k-means procedure \cite{Celebi_2013, Xiao2024ComprehensiveKMeans}. Let $\{\mathbf{h}_i\}_{i=1}^N$ denote the
$N$ tile embeddings used for concept learning, where each $\mathbf{h}_i \in \mathbb{R}^{d_h}$ is a $d_h$-dimensional feature
vector for tile $i$. We learn $K$ concept centroids $\{\boldsymbol{\mu}_k\}_{k=1}^K$ with $\boldsymbol{\mu}_c \in \mathbb{R}^{d_h}$
and a hard assignment function $c(i)\in\{1,\ldots,K\}$ that maps each tile $i$ to its nearest centroid by minimizing the
within-cluster sum of squared distances:
\begin{align}
\{\boldsymbol{\mu}_k\}_{k=1}^K,\ \{c(i)\}_{i=1}^N
&= \argmin_{\{\boldsymbol{\mu}_k\},\, c(i)}
\sum_{i=1}^N \left\|\mathbf{h}_i - \boldsymbol{\mu}_{c(i)}\right\|_2^2.
\label{eq:kmeans_rawh}
\end{align}
% \hao{(1) for each variable above explain what it means. (2) this equation merges both the assignment step and the center adjustment step, below this equation, you can also write down these two step separately. (3) it should be argmin }
% \sout{In practice, we initialize centroids with a reservoir-based minibatch K-Means pass and then run full Lloyd
% refinement to obtain stable concept centroids}
In practice, we initialize concept centroids in the $d_h$-dimensional $h$-space using a reservoir-based mini-batch K-means pass, followed by standard K-means clustering\cite{1056489} applied to all tile embeddings to obtain stable concept centroids. This variant treats all tiles equally during concept learning. We next introduce \emph{CLEAR-HPV (AW-$h$)}, which incorporates attention weights when forming concepts and provides a complementary view of the attention-structured $h$-space. 

\paragraph{CLEAR-HPV (AW-$h$).} To incorporate the MIL model's diagnostic relevance into concept learning, CLEAR-HPV (AW-$h$) uses an attention-weighted k-means objective.
Let $\{\mathbf{h}_i\}_{i=1}^N$ denote tile embeddings and let $\alpha_i \ge 0$ denote the corresponding attention weight
assigned by the MIL model (with $\sum_i \alpha_i > 0$). We learn $K$ centroids $\{\boldsymbol{\mu}_k\}_{k=1}^K$ and hard
assignments $c(i)\in\{1,\ldots,K\}$ by solving
\begin{align}
\{\boldsymbol{\mu}_k\}_{k=1}^K,\ \{c(i)\}_{i=1}^N
= \argmin_{\{\boldsymbol{\mu}_k\},\, c(i)}
\sum_{i=1}^N \alpha_i \left\|\mathbf{h}_i - \boldsymbol{\mu}_{c(i)}\right\|_2^2.
\label{eq:kmeans_awh}
\end{align}

During initialization and refinement, tile weights $\alpha_i$ act as sample weights, causing high-attention tiles to exert stronger influence on concept boundaries. This aligns the learned concepts with regions emphasized by the MIL model.

\paragraph{Dirichlet concepts.}
Building on ideas from probabilistic concept models such as PACE \cite{PACE}, we include a probabilistic concept model that learns uncertainty-aware importance score for each tile and each concept for the prediction. A probabilistic model like this is usually more robust and captures broader variability in morphologic patterns across slides.

\paragraph{Heatmap.} We evaluated a heatmap-based slide classifier that uses the MIL attention map as a quantitative decision signal. For each slide, tile-level attention scores are projected back onto the WSI to form a spatial heatmap. A slide-level prediction is obtained by computing a scalar “heatmap area” score, defined as the proportion of tiles whose attention exceeds a fixed threshold $\tau$ for the class of interest. The decision cutoff is selected on training data and applied to held-out test slides, yielding a transparent rule-based classifier grounded in attention-based evidence. 

\paragraph{Encoder concepts.} As a baseline, we evaluate an encoder-based concept discovery method that follows the same procedure as CLEAR-HPV (raw-$h$) but operates directly on encoder feature space rather than the MIL latent $h$-space. Let $\{\mathbf{z}_i\}_{i=1}^N$, with $\mathbf{z}_i \in \mathbb{R}^{d_e}$, denote the tile-level embeddings produced by the pretrained encoder; all subsequent steps are identical to CLEAR-HPV (raw-$h$). Because this baseline does not involve MIL attention, an attention-weighted (AW-$h$) variant is not applicable.

\paragraph{Choice of number of concepts ($K$).}  
We use $K=10$ as the main setting in our experiments. This choice was initially guided by elbow analysis~\cite{thorndike1953family} and further supported by evaluations across nearby values of $K$. These analyses show that predictive performance remains stable and that the overall structure of the discovered concepts is highly consistent across different values of $K$. In practice, changing $K$ mainly merges or splits similar concepts rather than producing entirely different patterns. We therefore do not interpret $K=10$ as a biologically unique optimum, but as a stable and interpretable choice for this study.

\subsection*{Concept-fraction Vectors in CLEAR-HPV}

As shown in~\figref{fig:main}(D) and \figref{fig:main}(G), our CLEAR-HPV provides both class-level and slide-level concept-fraction vectors to enable holistic interpretation (explanation) of MIL models' predictions. Below we describe details on how they are computed. 

\paragraph{Raw slide-level concept-fraction vectors.} Given a set of $K$ discovered concepts from CLEAR-HPV, we represent each whole slide image using a \emph{concept-fraction vector} that summarizes its morphologic composition. Specifically, for a slide with $N$ tiles, let $c(i)\in\{1,\dots,K\}$ denote the index of the concept assigned to tile $i\in \{1,2,\dots,N\}$. The slide-level concept-fraction vector $\mathbf{f}\in\mathbb{R}^K$ is defined as
\begin{align}
f_k = \frac{1}{N}\sum_{i=1}^N \mathbb{I}\big[c(i)=k\big], \quad k=1,\ldots,K, \label{eq:cfv_raw}
\end{align}
where $f_k$ is the $k$-th entry in $\mathbf{f}$, and $\mathbb{I}[\cdot]$ is the indicator function, equal to $1$ if tile $i$ is assigned to concept $k$ and $0$ otherwise. Each entry $f_k$ therefore represents the proportion of tiles in the slide assigned to concept $k$, and the vector $\mathbf{f}$ is normalized such that $\sum_{k=1}^K f_k = 1$.

\paragraph{Attention-weighted slide-level concept-fraction vectors.} For attention-weighted variants, each tile $i$ is additionally associated with an attention weight $\alpha_i$ produced by the MIL model, reflecting its contribution to the slide-level prediction. Correspondingly, the \emph{attention-weighted (AW) concept fraction vectors} are computed as
\begin{align}
f_k = \frac{\sum_{i=1}^N \alpha_i\,\mathbb{I}\big[c(i)=k\big]}{\sum_{i=1}^N \alpha_i}, \label{eq:cfv_aw}
\end{align}
where tiles receiving higher attention contribute more strongly to the slide-level representation. Attention weights are normalized within each slide to ensure comparability across slides of different sizes. 

These two variants of slide-level concept-fraction vectors are used as inputs to the concept-fraction classifier for quantitative evaluation.

\paragraph{Class-averaged concept-fraction vectors.} For interpretability analyses, we further compute class-averaged concept-fraction vectors by averaging slide-level concept-fraction vectors across all slides within a given cohort or clinical group (e.g., averaging over HPV-positive cases or HPV-negative cases).
Throughout the paper, we use the term concept-fraction vector to denote slide-level representations, and class-averaged concept-fraction vectors when these representations are averaged within a subgroup of a cohort (e.g., HPV-positive cases) for interpretability analyses.

\subsection*{Representative tiles per concept}
As shown in~\figref{fig:main}(E), to support qualitative interpretation of the discovered concepts, we identify representative tiles for each concept using an automated and reproducible ranking procedure. Tiles are ranked {by their Euclidean distance to the corresponding concept centroid
$\boldsymbol{\mu}_k \in \{\boldsymbol{\mu}_k\}_{k=1}^K$, as defined by \eqnref{eq:kmeans_rawh}}, 
and the top $M$ tiles are selected across slides for visualization. These representative tiles are used exclusively for qualitative interpretation and do not influence model training or evaluation.

\subsection*{Spatial concept maps}

\paragraph{Spatial concept maps.} 
As shown in~\figref{fig:main}(F) and \figref{fig:clustering}(A)(ii), to visualize the spatial organization of discovered concepts within WSIs, we generate spatial concept maps by projecting tile-level concept assignments back to their original spatial locations in the WSI. Each tile is assigned to a single concept using hard assignment, and tiles are colored according to their assigned concept, producing a map that illustrates how different morphologic concepts are distributed across the tissue section. 

\paragraph{High-attention spatial concept maps.} 
As shown in~\figref{fig:clustering}(A)(iii), when attention scores are available, we additionally generate a high-attention map by displaying only the top $M$ tiles ranked by MIL attention score, highlighting regions most strongly emphasized by the model.

\subsection*{Evaluation of concept-fraction vectors}
All concept-discovery methods were evaluated under a consistent 10-fold train/test
protocol applied to the $h$-space. For each fold, concepts were learned using only
training slides, and concept-fraction vectors were computed for both training and
held-out slides. Slide-level predictions derived from concept fractions were compared against the baseline MIL classifier.

\paragraph{Concept-fraction classifier.} 

Given a slide $\mathcal{S}$, let $\mathbf{f}(\mathcal{S}) = [f_1,\ldots,f_K]\in\mathbb{R}^K$ denote its slide-level concept-fraction vector as defined in~\eqnref{eq:cfv_raw} and \eqnref{eq:cfv_aw}, where $f_k$ is the proportion of tiles assigned to concept $k$ and $\sum_{k=1}^K f_k = 1$. We then define a simple, interpretable rule-based classifier that scores each slide by aggregating the fractions of concepts that are statistically associated with HPV-positive status in the training data. Specifically, for each concept $k$, we compare the distribution of $f_k$ between HPV-positive and HPV-negative training slides and designate concept $k$ as HPV-associated if its mean fraction is higher in the HPV-positive group. This yields a binary concept mask $\boldsymbol{\pi}\in\{0,1\}^K$, where $\pi_k=1$ indicates that concept $k$ is positively associated with HPV status based on the training set, and $\pi_k=0$ otherwise. 

The slide-level score is computed as
\begin{align*}
s_{\mathrm{rule}}(\mathcal{S})=\sum_{k=1}^K \pi_k\, f_k(\mathcal{S}),
\end{align*}
which represents the total fraction of tissue assigned to HPV-associated concepts.
A binary prediction is obtained by thresholding this score,
\begin{align*}
\hat{y}(\mathcal{S})=\mathbb{I}\big[s_{\mathrm{rule}}(\mathcal{S})\ge \tau\big],
\end{align*}
where $\hat{y}(\mathcal{S})=1$ denotes an HPV-positive prediction and $\hat{y}(\mathcal{S})=0$ denotes an HPV-negative prediction. The decision threshold $\tau$ is selected using the training folds and then applied to held-out test slides.

For comparison, we also train a logistic regression classifier directly using the same concept-fraction vector $\mathbf{f}(\mathcal{S})$ as input. Detailed performance metrics for HPV and survival prediction are reported in Supplementary Table S8 and Table S9.

\paragraph{Recovery score.} 

Our CLEAR-HPV is a post-hoc explanation framework. It is therefore important to evaluate whether the generated explanation indeed reflects the explained model's prediction. To measure how well CLEAR-HPV variants and other baselines recover MIL predictions, we computed a recovery score by forming a metric vector, 
\begin{align*}
\m = [ACC, AUC, Prec, Rec, Spec, F1],
\end{align*}
for each method (e.g., $\m_{CLEAR-HPV}$ for CLEAR-HPV and $\m_{Base}$ for a certain base model) and calculating its Euclidean distance $d$ to the base model (e.g., CLAM) in the same fold. The final recovery score (for each fold) is
\begin{align*}
s=\frac{1}{1+d}, \quad d = \|\m_{CLEAR-HPV} - \m_{Base}\|_2
\end{align*}

\begin{description}

    \item Supplemental Methods S1. Definition of class-dominance ratios and dominant-cluster metrics (Avg Ratio and Sum Ratio).
    
    \item Figure S1. Elbow analysis used to select the number of morphologic concepts.
    
    \item Table S1. Performance across different numbers of concepts ($K$) for CLEAR-HPV (AW-$h$) on TCGA-HNSCC.
    
    \item Table S2. Stability of discovered concepts across different $K$, measured by forward persistence (FP) and reverse fragmentation (RF).

    \item Table S3. Results in terms of complete evaluation metrics for concept discovery methods on TCGA-HNSCC.
    
    \item Table S4. Effect of attention head selection in MHMIL-IR on TCGA-HNSCC.

    \item Table S5. Encoder-dependence analysis using ResNet50 comparing encoder-space concepts and CLEAR-HPV.
     
    \item Table S6. Results in terms of complete evaluation metrics for cross-cohort generalization from TCGA-HNSCC to TCGA-CESC.
    
    \item Table S7. Results in terms of complete evaluation metrics for TCGA-HNSCC survival prediction using CLEAR-HPV concepts derived from an HPV-trained backbone.
    
    \item Table S8. Logistic regression classifier-based HPV prediction on TCGA-HNSCC using CLEAR-HPV.
    
    \item Table S9. Logistic regression classifier-based survival prediction on TCGA-HNSCC using \\CLEAR-HPV.
\end{description}

\newpage

%%%  At final submission, figure files MUST be 
%%%  provided separately as high-resolution image 
%%%  files. All of the panels for a figure should 
%%%  be in the same file. Figures should have 
%%%  clear labels/file names (Figure 1, Figure 2, 
%%%  etc.). 

%%%  Figure titles and legends should be placed 
%%%  at the end of the main text. You do not 
%%%  need to place the figures, nor their titles 
%%%  and legends, within the main text. While 
%%%  typesetting your article, our team will 
%%%  place each figure in the best location 
%%%  based on the final layout and on your 
%%%  figure citations, e.g., (Figures 1A and 1B). 

%%%  Please review the figure guidelines before 
%%%  submitting your final materials: 
%%%  https://www.cell.com/figureguidelines.

% \section*{MAIN FIGURE TITLES AND LEGENDS}

% % Figure 1 — MAIN
% \noindent\includegraphics[width=\linewidth]{main.png}
% \subsection*{Figure \refstepcounter{figure}\thefigure.\label{fig:main} Attention-weighted \(h\)-space clustering yields interpretable concept profiles that preserve predictive performance}
% Pipeline: feature extraction, MIL backbone, attention-weighted \(h\)-space, clustering, slide-level concept proportions, and interpretable map.

\vspace{1.5em}

\newpage

%%%  REFERENCES: As of 2023, all Cell Press journals 
%%%  use Numbered (AMA) style. We recommend placing 
%%%  your references in the included "references.bib" 
%%%  file.

\bibliography{references}

@article{Campanella2019Clinical,
  author    = {Campanella, Giacomo and Hanna, Michael G. and Geneslaw, Lisa and Miraflor, Albert and Werneck Krauss Silva, Victor and Busam, Klaus J. and Brogi, Edi and Reuter, Victor E. and Klimstra, David S. and Fuchs, Thomas J.},
  title     = {Clinical-grade computational pathology using weakly supervised deep learning on whole slide images},
  journal   = {Nature Medicine},
  volume    = {25},
  pages     = {1301--1309},
  year      = {2019},
  doi       = {10.1038/s41591-019-0508-1}
}

@article{de2020global,
  title={Global burden of cancer attributable to infections in 2018: a worldwide incidence analysis},
  author={de Martel, Catherine and Georges, Damien and Bray, Freddie and Ferlay, Jacques and Clifford, Gary M},
  journal={The Lancet global health},
  volume={8},
  number={2},
  pages={180--190},
  year={2020}
}

@article{Coudray2018NSCLC,
  author    = {Coudray, Nicolas and Ocampo, Paola S. and Sakellaropoulos, Theodore and Narula, Navneet and Snuderl, Matija and Feny{\"o}, David and Moreira, Andre L. and Razavian, Narges and Tsirigos, Aristotelis},
  title     = {Classification and mutation prediction from non--small cell lung cancer histopathology images using deep learning},
  journal   = {Nature Medicine},
  volume    = {24},
  pages     = {1559--1567},
  year      = {2018},
  doi       = {10.1038/s41591-018-0177-5}
}

@article{Liu2020_AI_NodalMetastasisDetection,
  author       = {Liu, Y. and Kohlberger, T. and Norouzi, M. and Dahl, G. E. and Metz, L. and Shoemaker, J. and MacDonald, R. and Scott, C. and Nelson, P. Q. and Hipp, J. D. and Corrado, G. S. and Dean, J. and MacMahon, H. and Madabhushi, A. and Stumpe, M. C.},
  title        = {Artificial intelligence–based breast cancer nodal metastasis detection: Insights into the black box for pathologists},
  journal      = {JAMA},
  year         = {2020},
  volume       = {323},
  number       = {4},
  pages        = {315--325},
  doi          = {10.1001/jama.2019.2625}
}

@article{samek2017explainable,
  title={Explainable Artificial Intelligence: Understanding, Visualizing and Interpreting Deep Learning Models},
  author={Samek, Wojciech and Wiegand, Thomas and M{\"u}ller, Klaus-Robert},
  journal={IEEE Signal Processing Magazine},
  volume={34},
  number={6},
  pages={26--41},
  year={2017},
  publisher={IEEE}
}

@article{arrieta2020explainable,
  title={Explainable Artificial Intelligence (XAI): Concepts, Taxonomies, Opportunities and Challenges toward Responsible AI},
  author={Arrieta, Alejandro Barredo and D{\'\i}az-Rodr{\'\i}guez, Natalia and Del Ser, Javier and Bennetot, Adrien and Tabik, Siham and Barbado, Alberto and Garc{\'\i}a, Salvador and Gil-L{\'o}pez, Sergio and Molina, Daniel and Benjamins, Richard and others},
  journal={Information Fusion},
  volume={58},
  pages={82--115},
  year={2020},
  publisher={Elsevier}
}

@article{tjoa2020survey,
  title={A Survey on Explainable Artificial Intelligence (XAI): Toward Medical XAI},
  author={Tjoa, Erico and Guan, Cuntai},
  journal={Computer Methods and Programs in Biomedicine},
  volume={200},
  pages={105009},
  year={2021},
  publisher={Elsevier}
}

@article{ITW:2018,
  title={Attention-based Deep Multiple Instance Learning},
  author={Ilse, Maximilian and Tomczak, Jakub M and Welling, Max},
  journal={arXiv preprint arXiv:1802.04712},
  year={2018}
}

@article{lu2021clam,
  title={Data-efficient and Weakly Supervised Computational Pathology on Whole-slide Images},
  author={Lu, Ming Y. and Williamson, Drew F. K. and Chen, Tiffany Y. and Chen, Rui and Barbieri, Mauro and Mahmood, Faisal},
  journal={Nature Biomedical Engineering},
  volume={5},
  pages={555--570},
  year={2021},
  doi={10.1038/s41551-020-00682-w}
}

@article{shao2021transmil,
  title={Transmil: Transformer based correlated multiple instance learning for whole slide image classification},
  author={Shao, Zhuchen and Bian, Hao and Chen, Yang and Wang, Yifeng and Zhang, Jian and Ji, Xiangyang and others},
  journal={Advances in Neural Information Processing Systems},
  volume={34},
  pages={2136--2147},
  year={2021}
}

@article{weinstein2013cancer,
  title={The Cancer Genome Atlas Pan-Cancer Analysis Project},
  author={Weinstein, John N and others},
  journal={Nature Genetics},
  volume={45},
  number={10},
  pages={1113--1120},
  year={2013},
  publisher={Nature Publishing Group}
}

@article{edwards2015cptac,
  title={The Cancer Proteome Atlas (CPTAC): a community resource for proteogenomic cancer research},
  author={Edwards, Nathan J and others},
  journal={Journal of Proteome Research},
  volume={14},
  number={6},
  pages={2707--2713},
  year={2015}
}

@article{TCGA_CESC_2017,
  author    = {The Cancer Genome Atlas Research Network},
  title     = {Integrated genomic and molecular characterization of cervical cancer},
  journal   = {Nature},
  volume    = {543},
  pages     = {378--384},
  year      = {2017},
  doi       = {10.1038/nature21386}
}

@article{chen2024uni,
  title={Towards a General-Purpose Foundation Model for Computational Pathology},
  author={Chen, Richard J and Ding, Tong and Lu, Ming Y and Williamson, Drew FK and Jaume, Guillaume and Chen, Bowen and Zhang, Andrew and Shao, Daniel and Song, Andrew H and Shaban, Muhammad and others},
  journal={Nature Medicine},
  publisher={Nature Publishing Group},
  year={2024}
}

@article{xu2024gigapath,
  title={A whole-slide foundation model for digital pathology from real-world data},
  author={Xu, Hanwen and Usuyama, Naoto and Bagga, Jaspreet and Zhang, Sheng and Rao, Rajesh and Naumann, Tristan and Wong, Cliff and Gero, Zelalem and González, Javier and Gu, Yu and Xu, Yanbo and Wei, Mu and Wang, Wenhui and Ma, Shuming and Wei, Furu and Yang, Jianwei and Li, Chunyuan and Gao, Jianfeng and Rosemon, Jaylen and Bower, Tucker and Lee, Soohee and Weerasinghe, Roshanthi and Wright, Bill J. and Robicsek, Ari and Piening, Brian and Bifulco, Carlo and Wang, Sheng and Poon, Hoifung},
  journal={Nature},
  year={2024},
  publisher={Nature Publishing Group UK London}
}

@article{Kather2019MSI,
  title={Deep learning can predict microsatellite instability directly from histology in gastrointestinal cancer},
  author={Kather, Jakob N and Pearson, Alexander T and Halama, Niels and J{\"a}ger, Dirk and Krause, Jos{\'e} and Loosen, Stefanie H and Marx, Alexander and Foersch, Sabine and D{\"u}sterhoft, Ansgar and van den Heuvel, Thomas and {et~al.}},
  journal={Nature Medicine},
  volume={25},
  pages={1054--1056},
  year={2019},
  doi={10.1038/s41591-019-0462-y}
}

@article{SelfSupervisedHisto,
title = {Self supervised contrastive learning for digital histopathology},
journal = {Machine Learning with Applications},
volume = {7},
pages = {100198},
year = {2022},
issn = {2666-8270},
doi = {https://doi.org/10.1016/j.mlwa.2021.100198},
url = {https://www.sciencedirect.com/science/article/pii/S2666827021000992},
author = {Ozan Ciga and Tony Xu and Anne Louise Martel}
}

@article{Mobadersany2018PNAS,
  title={Predicting cancer outcomes from histology and genomics using convolutional networks},
  author={Mobadersany, Pooya and Yousefi, Sahand and Amgad, Mohamed and Gutman, David A. and Barnholtz-Sloan, Jill S. and Vel{\'a}zquez Vega, Jorge E. and Brat, Daniel J. and Cooper, Lee A. D.},
  journal={Proceedings of the National Academy of Sciences},
  volume={115},
  number={13},
  pages={E2970--E2979},
  year={2018},
  month={Mar},
  doi={10.1073/pnas.1717139115},
  pmid={29531073},
  pmcid={PMC5879673}
}

@article{Xiao2024ComprehensiveKMeans,
  title={Comprehensive K-Means Clustering},
  author={Xiao, Ethan},
  journal={Journal of Computer and Communications},
  volume={12},
  number={3},
  pages={},
  year={2024},
  month={March 26},
  doi={}
}

@article{Celebi_2013,
   title={A comparative study of efficient initialization methods for the k-means clustering algorithm},
   volume={40},
   ISSN={0957-4174},
   url={http://dx.doi.org/10.1016/j.eswa.2012.07.021},
   DOI={10.1016/j.eswa.2012.07.021},
   number={1},
   journal={Expert Systems with Applications},
   publisher={Elsevier BV},
   author={Celebi, M. Emre and Kingravi, Hassan A. and Vela, Patricio A.},
   year={2013},
   month=jan, pages={200–210} }

@misc{bengio2014representationlearningreviewnew,
      title={Representation Learning: A Review and New Perspectives}, 
      author={Yoshua Bengio and Aaron Courville and Pascal Vincent},
      year={2014},
      eprint={1206.5538},
      archivePrefix={arXiv},
      primaryClass={cs.LG},
      url={https://arxiv.org/abs/1206.5538}, 
}

@misc{ghorbani2019automaticconceptbasedexplanations,
      title={Towards Automatic Concept-based Explanations}, 
      author={Amirata Ghorbani and James Wexler and James Zou and Been Kim},
      year={2019},
      eprint={1902.03129},
      archivePrefix={arXiv},
      primaryClass={stat.ML},
      url={https://arxiv.org/abs/1902.03129}, 
}

@misc{caron2021emergingpropertiesselfsupervisedvision,
      title={Emerging Properties in Self-Supervised Vision Transformers}, 
      author={Mathilde Caron and Hugo Touvron and Ishan Misra and Hervé Jégou and Julien Mairal and Piotr Bojanowski and Armand Joulin},
      year={2021},
      eprint={2104.14294},
      archivePrefix={arXiv},
      primaryClass={cs.CV},
      url={https://arxiv.org/abs/2104.14294}, 
}

@misc{koh2020conceptbottleneckmodels,
      title={Concept Bottleneck Models}, 
      author={Pang Wei Koh and Thao Nguyen and Yew Siang Tang and Stephen Mussmann and Emma Pierson and Been Kim and Percy Liang},
      year={2020},
      eprint={2007.04612},
      archivePrefix={arXiv},
      primaryClass={cs.LG},
      url={https://arxiv.org/abs/2007.04612}, 
}

@article{Selvaraju_2019,
   title={Grad-CAM: Visual Explanations from Deep Networks via Gradient-Based Localization},
   volume={128},
   ISSN={1573-1405},
   url={http://dx.doi.org/10.1007/s11263-019-01228-7},
   DOI={10.1007/s11263-019-01228-7},
   number={2},
   journal={International Journal of Computer Vision},
   publisher={Springer Science and Business Media LLC},
   author={Selvaraju, Ramprasaath R. and Cogswell, Michael and Das, Abhishek and Vedantam, Ramakrishna and Parikh, Devi and Batra, Dhruv},
   year={2019},
   month=oct, pages={336–359} }

@article{Ang2010NEJM,
  author    = {Ang, Kevin K. and Harris, Jay and Wheeler, Robert and Weber, Randal and Rosenthal, David I. and Nguyen-T{\^a}n, P. F. and Westra, William H. and Chung, Christine H. and Jordan, R. Clay and Lu, Chaomei and Kim, Hyejin and Axelrod, Rebecca and Silverman, Carole C. and Redmond, Kimberly P. and Gillison, Maura L.},
  title     = {Human papillomavirus and survival of patients with oropharyngeal cancer},
  journal   = {The New England Journal of Medicine},
  volume    = {363},
  number    = {1},
  pages     = {24--35},
  year      = {2010},
  month     = jul,
  doi       = {10.1056/NEJMoa0912217},
  pmid      = {20530316},
  pmcid     = {PMC2943767}
}

@article{Gillison2000JNCI,
  author    = {Gillison, Maura L. and Koch, William M. and Capone, Ralph B. and Spafford, Michael and Westra, William H. and Wu, Ling and Zahurak, Mark L. and Daniel, Robert W. and Viglione, Michael and Symer, Daniel E. and Shah, Keerti V. and Sidransky, David},
  title     = {Evidence for a causal association between human papillomavirus and a subset of head and neck cancers},
  journal   = {Journal of the National Cancer Institute},
  volume    = {92},
  number    = {9},
  pages     = {709--720},
  year      = {2000},
  month     = may,
  doi       = {10.1093/jnci/92.9.709},
  pmid      = {10793107}
}

@misc{goyal2020explainingclassifierscausalconcept,
      title={Explaining Classifiers with Causal Concept Effect (CaCE)}, 
      author={Yash Goyal and Amir Feder and Uri Shalit and Been Kim},
      year={2020},
      eprint={1907.07165},
      archivePrefix={arXiv},
      primaryClass={cs.LG},
      url={https://arxiv.org/abs/1907.07165}, 
}

@article{Lu2024NatMed,
  author    = {Lu, Ming Y. and Chen, Bowen and Williamson, Drew F. K. and Chen, Richard J. and Liang, Isaac and Ding, Tian and Jaume, Guillaume and Odintsov, Igor and Le, Long Phi and Gerber, Georg and Parwani, Anil V. and Zhang, Ann and Mahmood, Faisal},
  title     = {A visual-language foundation model for computational pathology},
  journal   = {Nature Medicine},
  volume    = {30},
  number    = {3},
  pages     = {863--874},
  year      = {2024},
  month     = mar,
  doi       = {10.1038/s41591-024-02856-4},
  pmid      = {38504017},
  pmcid     = {PMC11384335}
}

@misc{nicke2025tissueconceptsv2supervised,
      title={Tissue Concepts v2: A Supervised Foundation Model For Whole Slide Images}, 
      author={Till Nicke and Daniela Schacherer and Jan Raphael Schäfer and Natalia Artysh and Antje Prasse and André Homeyer and Andrea Schenk and Henning Höfener and Johannes Lotz},
      year={2025},
      eprint={2507.05742},
      archivePrefix={arXiv},
      primaryClass={eess.IV},
      url={https://arxiv.org/abs/2507.05742}, 
}

@article{Stacke2021DomainShiftHistopathology,
  title        = {Measuring Domain Shift for Deep Learning in Histopathology},
  author       = {Stacke, Kristofer and Eilertsen, Gabriel and Unger, Jonas and Lundstr{\"o}m, Claes},
  journal      = {IEEE Journal of Biomedical and Health Informatics},
  year         = {2021},
  volume       = {25},
  number       = {2},
  pages        = {325--336},
  doi          = {10.1109/JBHI.2020.3032060},
  pmid         = {33085623}
}

@article{Zarella2019PracticalWSI,
  title        = {A Practical Guide to Whole Slide Imaging: A White Paper From the Digital Pathology Association},
  author       = {Zarella, Mark D. and Bowman, Deborah and Aeffner, Famke and Farahani, Navid and Xthona, Andrew and Absar, Syed F. and Parwani, Anil and Bui, Marilyn and Hartman, Douglas J.},
  journal      = {Archives of Pathology \& Laboratory Medicine},
  year         = {2019},
  volume       = {143},
  number       = {2},
  pages        = {222--234},
  doi          = {10.5858/arpa.2018-0343-RA},
  pmid         = {30307746}
}

@article{Linton2013BasaloidSCC,
  author       = {Linton, O. R. and Moore, M. G. and Brigance, J. S. and Gordon, C. A. and Summerlin, D. J. and McDonald, M. W.},
  title        = {Prognostic significance of basaloid squamous cell carcinoma in head and neck cancer},
  journal      = {JAMA Otolaryngology--Head \& Neck Surgery},
  year         = {2013},
  volume       = {139},
  number       = {12},
  pages        = {1306--1311},
  doi          = {10.1001/jamaoto.2013.5308},
  pmid         = {24158536}
}

@article{Lewis2018HPVTestingGuideline,
  author       = {Lewis, J. S. Jr and Beadle, B. and Bishop, J. A. and Chernock, R. D. and Colasacco, C. and Lacchetti, C. and Moncur, J. T. and Rocco, J. W. and Schwartz, M. R. and Seethala, R. R. and Thomas, N. E. and Westra, W. H. and Faquin, W. C.},
  title        = {Human Papillomavirus Testing in Head and Neck Carcinomas: Guideline From the College of American Pathologists},
  journal      = {Archives of Pathology \& Laboratory Medicine},
  year         = {2018},
  volume       = {142},
  number       = {5},
  pages        = {559--597},
  month        = may,
  doi          = {10.5858/arpa.2017-0286-CP},
  pmid         = {29251996}
}

@article{Marur2008HNCChangingEpidemiology,
  author       = {Marur, S. and Forastiere, A. A.},
  title        = {Head and neck cancer: changing epidemiology, diagnosis, and treatment},
  journal      = {Mayo Clinic Proceedings},
  year         = {2008},
  volume       = {83},
  number       = {4},
  pages        = {489--501},
  month        = apr,
  doi          = {10.4065/83.4.489},
  pmid         = {18380996},
  note         = {Erratum in Mayo Clin Proc. 2008 May;83(5):604}
}

@article{Shah2014SCCVariantsReview,
  author       = {Shah, A. A. and Jeffus, S. K. and Stelow, E. B.},
  title        = {Squamous cell carcinoma variants of the upper aerodigestive tract: a comprehensive review with a focus on genetic alterations},
  journal      = {Archives of Pathology \& Laboratory Medicine},
  year         = {2014},
  volume       = {138},
  number       = {6},
  pages        = {731--744},
  month        = jun,
  doi          = {10.5858/arpa.2013-0070-RA},
  pmid         = {24878013}
}

@inproceedings{PACE,
  title={Probabilistic Conceptual Explainers: Trustworthy Conceptual Explanations for Vision Foundation Models},
  author={Hengyi Wang and
          Shiwei Tan and
          Hao Wang},
  booktitle={International Conference on Machine Learning},
  year={2024}
}

@article{thorndike1953family,
  title={Who belongs in the family?},
  author={Thorndike, Robert L.},
  journal={Psychometrika},
  volume={18},
  number={4},
  pages={267--276},
  year={1953},
  publisher={Springer}
}

@ARTICLE{1056489,
  author={Lloyd, S.},
  journal={IEEE Transactions on Information Theory}, 
  title={Least squares quantization in PCM}, 
  year={1982},
  volume={28},
  number={2},
  pages={129-137},
  doi={10.1109/TIT.1982.1056489}}

@ARTICLE{9359803,
  author={Haghighi, Fatemeh and Taher, Mohammad Reza Hosseinzadeh and Zhou, Zongwei and Gotway, Michael B. and Liang, Jianming},
  journal={IEEE Transactions on Medical Imaging}, 
  title={Transferable Visual Words: Exploiting the Semantics of Anatomical Patterns for Self-Supervised Learning}, 
  year={2021},
  volume={40},
  number={10},
  pages={2857-2868},
  doi={10.1109/TMI.2021.3060634}}

@inproceedings{xia-etal-2025-hgclip,
    title = "{HGCLIP}: Exploring Vision-Language Models with Graph Representations for Hierarchical Understanding",
    author = "Xia, Peng  and
      Yu, Xingtong  and
      Hu, Ming  and
      Ju, Lie  and
      Wang, Zhiyong  and
      Duan, Peibo  and
      Ge, Zongyuan",
    editor = "Rambow, Owen  and
      Wanner, Leo  and
      Apidianaki, Marianna  and
      Al-Khalifa, Hend  and
      Eugenio, Barbara Di  and
      Schockaert, Steven",
    booktitle = "Proceedings of the 31st International Conference on Computational Linguistics",
    month = jan,
    year = "2025",
    address = "Abu Dhabi, UAE",
    publisher = "Association for Computational Linguistics",
    url = "https://aclanthology.org/2025.coling-main.19/",
    pages = "269--280"
}

@misc{gu2025radalignadvancingradiologyreport,
      title={RadAlign: Advancing Radiology Report Generation with Vision-Language Concept Alignment}, 
      author={Difei Gu and Yunhe Gao and Yang Zhou and Mu Zhou and Dimitris Metaxas},
      year={2025},
      eprint={2501.07525},
      archivePrefix={arXiv},
      primaryClass={cs.CV},
      url={https://arxiv.org/abs/2501.07525}, 
}

\bigskip

%%%  In your References, please include only articles 
%%%  that are published (online publication and 
%%%  preprint servers are OK). Unpublished data, 
%%%  submitted and/or accepted manuscripts, abstracts, 
%%%  and personal communications should be cited within 
%%%  the text only ("unpublished data," "data not 
%%%  shown," "Alice Smith, personal communication") 
%%%  and not included in the references list. Personal 
%%%  communication should be documented by a letter 
%%%  of permission. Whenever possible, please make 
%%%  sure your .bib file has the complete author lists 
%%%  for each item (at minimum, the first 11 authors 
%%%  listed). 

%%%  ADDITIONAL MANUSCRIPT COMPONENTS:

%%%  Depending on the journal and the article 
%%%  type, you may be asked to upload the 
%%%  following as separate files: graphical 
%%%  abstract, highlights, eTOC blurb (In 
%%%  Brief), and/or other article components 
%%%  such as a "bigger picture" statement. 
%%%  These items are typically not required for 
%%%  initial submissions. Please refer to the 
%%%  journal's website, your acceptance 
%%%  letter, and/or the Final Files 
%%%  Requirements checklist to see if any of
%%%  these items are required (at any stage).

\end{document}